\pdfoutput=1
\documentclass[11pt]{article}
\usepackage[most]{tcolorbox}
\usepackage{amsmath}
\usepackage{dsfont}
\usepackage{acl}

\usepackage{times}
\usepackage{latexsym}
\usepackage{pifont}

\usepackage[T1]{fontenc}

\usepackage[utf8]{inputenc}

\usepackage{microtype}
\usepackage{hyperref}
\usepackage{inconsolata}

\usepackage{graphicx}
\usepackage{xspace}
\newcommand{\method}{\textsc{MMUnlearner}\xspace}

\usepackage{hyperref}
\usepackage{url}
\usepackage{hyperref}
\usepackage[most]{tcolorbox}
\usepackage{wrapfig}
\usepackage{graphicx,xcolor,float}
\usepackage{subcaption}
\usepackage{threeparttable}
\usepackage{algorithm}
\usepackage{pifont}
\usepackage[noend]{algorithmic}
\usepackage{colortbl}
\usepackage{color}
\usepackage{multirow}
\usepackage{tabularx}
\usepackage{float}
\usepackage{graphicx}
\usepackage{booktabs}
\usepackage{arydshln}
\usepackage{enumitem}
\usepackage{wrapfig}
\usepackage{caption}
\usepackage{graphicx}
\usepackage{makecell}
\usepackage{float} 
\usepackage{makecell}
\usepackage{tabularx}
\usepackage{twemojis}
\usepackage{amssymb,mathrsfs,amsmath}

\usepackage{pifont}
\usepackage[export]{adjustbox}
\usepackage{xcolor}
\usepackage{xspace}
\usepackage{shadowtext}
\usepackage{anyfontsize}
\usepackage{array}
\usepackage{caption}
\usepackage{subcaption}
\newcommand\sbullet[1][.5]{\mathbin{\vcenter{\hbox{\scalebox{#1}{$\bullet$}}}}}
\usepackage{xcolor}
\usepackage{graphicx}
\usepackage{pifont}
\usepackage{multirow} 
\usepackage{xcolor,colortbl}
\usepackage{times}
\usepackage{latexsym}
\usepackage{arydshln} 
\usepackage[acl,subfig]{definition}
\usepackage[detect-none]{siunitx}

\hypersetup{
    colorlinks=true,
    linkcolor=red,
    citecolor=cyan,
    filecolor=magenta,      
    urlcolor=magenta,
    }

\definecolor{bittersweet}{rgb}{1.0, 0.44, 0.37}
\definecolor{mygreen}{rgb}{0.29, 0.7, 0.48}
\definecolor{my_green}{RGB}{51,102,0}
\definecolor{my_yellow}{RGB}{255,165,0}
\definecolor{my_red}{RGB}{204, 0, 0}

\usepackage{pifont}% http://ctan.org/pkg/pifont for xmark and cmark
\definecolor{demphcolor}{RGB}{144,144,144}

\definecolor{mygray}{gray}{0.4}
\hypersetup{
    colorlinks=true,
    linkcolor=red,
    citecolor=cyan,
    filecolor=magenta,      
    urlcolor=magenta,
    }
\usepackage{xcolor} % 如果你想为雪花上色
\usepackage{xcolor}
\usepackage[font=small,labelfont=bf]{caption}  % set global caption size
\usepackage{makecell}
\usepackage{tabulary}
\definecolor{ada_green}{rgb}{0,205,205}
\definecolor{glt_red}{rgb}{109,205,255}

\definecolor{backred}{RGB}{255, 190, 190}
\definecolor{backblue}{RGB}{210, 230, 250}
\definecolor{backgrey}{RGB}{220, 220, 220}

\definecolor{mygreen}{RGB}{184, 213, 118}
\definecolor{myred}{RGB}{215, 6, 84}
\definecolor{myblue}{RGB}{41, 115, 178}
\definecolor{shadecolor}{RGB}{237,237,237}

\makeatletter
\newcommand{\printfnsymbol}[1]{%
  \textsuperscript{\@fnsymbol{#1}}%
}
\makeatother
\title{\method: Reformulating Multimodal Machine Unlearning \\ in the Era of Multimodal Large Language Models}

\author{Jiahao~Huo\textsuperscript{\rm 1,3} \footnotemark[2],
  Yibo~Yan\textsuperscript{\rm 1,2} \footnotemark[2],\\
  \textbf{Xu~Zheng}$^{1,2}$,
  \textbf{Yuanhuiyi~Lyu}$^{1,2}$,
  \textbf{Xin~Zou}$^{1}$,
  \textbf{Zhihua~Wei}$^{3}$,
  \textbf{Xuming~Hu}\textsuperscript{\rm 1,2 *} \\
  \fontsize{9.0pt}{\baselineskip}\selectfont $^{1}$ The Hong Kong University of Science and Technology (Guangzhou) \\ 
  \fontsize{9.0pt}{\baselineskip}\selectfont $^{2}$ The Hong Kong University of Science and Technology, \fontsize{9.0pt}{\baselineskip}\selectfont $^{3}$ Tongji University \\
   \fontsize{9.0pt}{\baselineskip}\selectfont\texttt{\{jiahaohuotj, yanyibo70\}@gmail.com}, \texttt{\{xuminghu\}@hkust-gz.edu.cn}
}
\begin{document}
\maketitle
\renewcommand{\thefootnote}{\fnsymbol{footnote}}
\footnotetext[2]{Equal contribution.}
\footnotetext[1]{Corresponding author.}
\renewcommand{\thefootnote}{\arabic{footnote}}
\begin{abstract} 
Recent progress in Machine Unlearning (MU) has introduced solutions for the selective removal of private or sensitive information encoded within deep neural networks. Nonetheless, \textit{MU for Multimodal Large Language Models (MLLMs) remains in its nascent phase}. Therefore, we propose to \textbf{reformulate the task of multimodal MU in the era of MLLMs}, which aims to erase only the visual patterns associated with a given entity while preserving the corresponding textual knowledge encoded within the original parameters of the language model backbone. 
Furthermore, we develop \textbf{a novel geometry-constrained gradient ascent method \method}. It updates the weights of MLLMs with a weight saliency map jointly restricted by the remaining concepts and textual knowledge during unlearning, thereby preserving parameters essential for non-target knowledge. 
% It updates the weights of MLLMs within a manifold jointly rendered by the remaining geometry and textual knowledge during unlearning, thereby preserving parameters essential for non-target knowledge. 
% Our approach significantly erases targeted visual information while retains the intrinsic knowledge of language models. 
Extensive experiments demonstrate that \method surpasses baselines that finetuning MLLMs with VQA data directly through Gradient Ascent (GA) or Negative Preference Optimization (NPO), across all evaluation dimensions. Our code can be found in this \href{https://github.com/Z1zs/MMUnlearner}{URL}
\end{abstract}

\section{Introduction}
Multimodal Large Language Models (MLLMs) achieved remarkable performance on various multimodal applications~\cite{li2025benchmark,zou2025deep,yan2024survey,yan2024errorradar,dang2024explainable}. A common framework of MLLMs, which projects the visual embeddings extracted from pre-trained vision encoder into the representation space of language models with a projector, has enabled LLM backbone to understand visual inputs and preserve their powerful reasoning and generation potential~\cite{liu2024llava,yan2025position,huo2024mmneuron}. 
However, The rapid development of MLLM is also accompanied by safety concerns such as personal privacy~\cite{pi2024mmprivacy} and copyright infringement~\cite{li2024digger}. Retraining the models from scratch to exclude the risky knowledge is resource-intensive and practically untenable due to the inaccessible pre-training data~\cite{bourtoule2021mu,si2023llmu_survey}. Hence, Machine Unlearning (MU) can serve as a feasible solution to forget specific knowledge embedded within pre-trained models~\cite{blanco2025llm_mu}.\par

\begin{figure}[!t]
    \centering
    \includegraphics[width=\linewidth,scale=1.00]{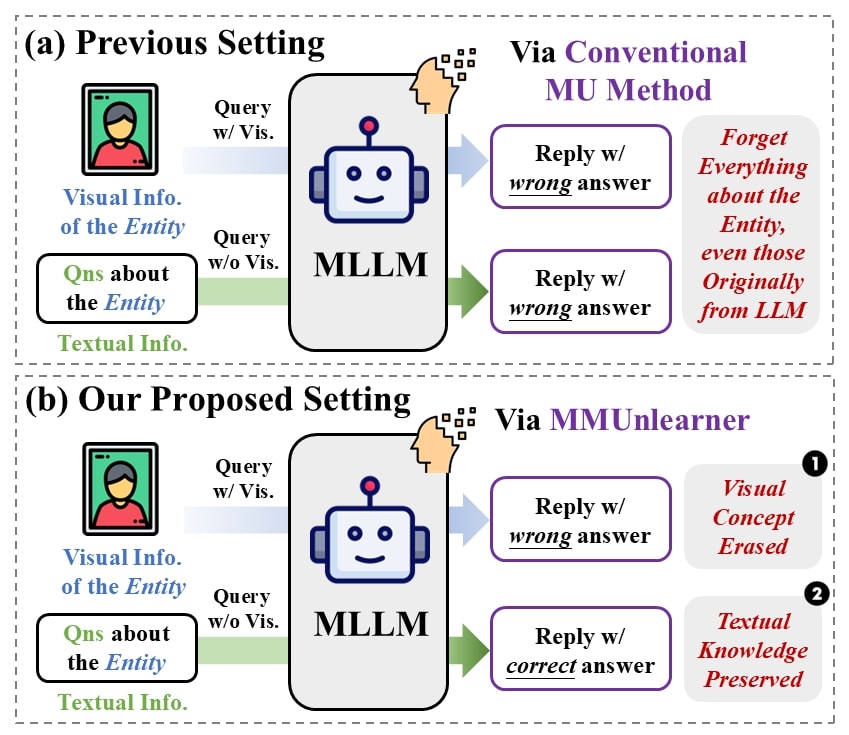}
    \caption{Comparison between the previous setting (a) and our proposed one (b) for multimodal machine unlearning.}
    \label{fig:paradigm_comparison}
    \vspace{-4mm}
\end{figure}

\begin{figure*}[thp]
    \centering
    \includegraphics[width=0.98\textwidth]{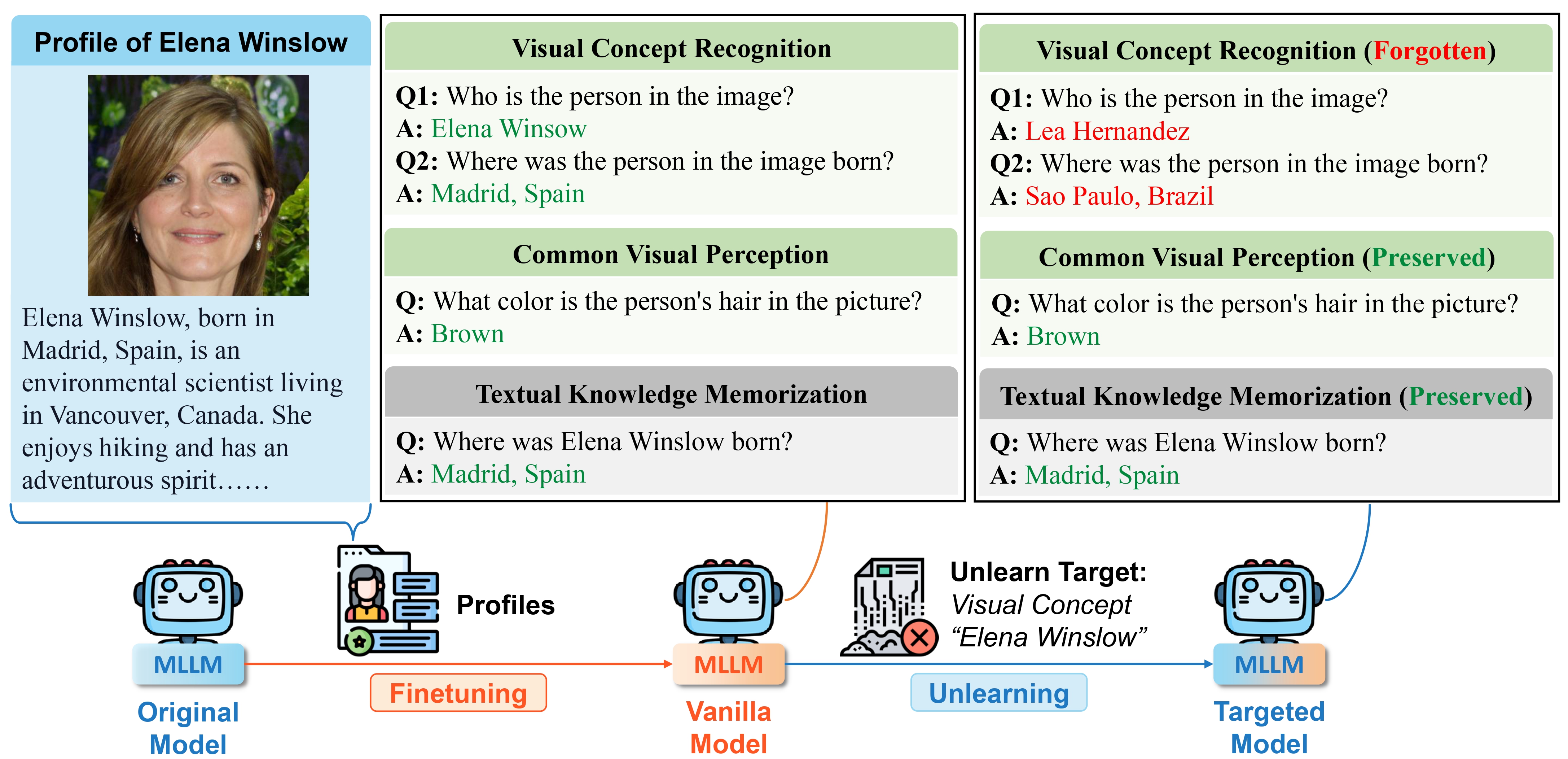}
    \caption{The framework of our reformulated \emph{Multimodal Machine Unlearning}. Different from LLM-based unlearning setting, it emphasizes the accurate removal of specific vision patterns of targeted concepts and the preservation of textual knowledge.}
    \label{fig:framework}
\end{figure*}

Nevertheless, \textit{MU on MLLMs is still in its nascent phase, with limited approaches and benchmarks available}. For example, Single Image Unlearning (SIU) first explores the MU of MLLMs, aiming to erase visual patterns in MLLMs on real-world entities, but it needs to reconstruct multifaceted fine-tuning data for forgetting~\cite{li2024siu}. Besides, MLLMU-Bench evaluates the performance of MU methods designed for LLMs on fictional personal profiles~\cite{liu2024mllmubench}; CLEAR adds visual images to pure-text LLM unlearning benchmark TOFU through Photomaker \cite{li2024photomaker}, a diffusion model adapted for customized realistic human~\cite{dontsov2024clear}. As shown in Figure \ref{fig:paradigm_comparison} (a), the aforementioned works just transfer LLM-based MU methods to MLLMs via fine-tuning on VQA data, and neglect the unique difficulty for MLLM-specific MU.\par

Therefore, we propose to \textbf{reformulate the task of multimodal MU} in the age of MLLMs, as illustrated in Figure \ref{fig:paradigm_comparison} (b). Unlike text-only LLMs, the knowledge embedded in MLLMs extends beyond textual factual knowledge within their LLM module, which includes learned visual attributes associated with various concepts~\cite{cohen2024performance,yu2024understanding}. Given this fundamental difference, we define the objective of MLLM-based MU as the selective removal of visual patterns linked to a specific entity, while preserving the corresponding textual knowledge within the LLM backbone, as illustrated in Figure \ref{fig:framework}. Considering that existing benchmarks have largely overlooked this crucial distinction, we aim to address this gap and ensure that multimodal MU methods can focus on the unique characteristics of MLLMs. 

To erase the memorized visual representation while preserving corresponding factual knowledge within MLLMs, we propose \textbf{\method, a geometry-constrained gradient ascent MU method to update the parameters for targeted visual patterns.} Motivated by the selective unlearning paradigm for visual networks and diffusion model~\cite{fan2023salun,huang2024remain}, we further extend it to MLLMs, with an appropriate saliency map (depicted by Fisher matrix in parameter space) designed for each module. Extensive experiment show that applying LLM-based MU methods to MLLMs with VQA data adjusts textual factual knowledge solely, whereas \method can efficiently remove the visual patterns while maintaining factual knowledge. Our findings offer valuable insights into multimodal intelligence in the era of Artificial General Intelligence (AGI).

Our contributions can be summarized as follows:

\ding{182} We are \textbf{the first to formulate the setting of Multimodal Machine Unlearning based on the characteristics of MLLM architecture during unlearning and evaluation}. Our focus is to erase the memorized visual representation while preserving corresponding factual knowledge.

\ding{183} We propose \textbf{a new weight saliency-based unlearning method, \method, to selectively update the parameter of MLLMs}, displaying superior performance in visual concepts erasing as well as preserving untargeted visual concepts and textual knowledge under the same setting.

\ding{184} We conduct \textbf{extensive experiments} on representative MLLMs and carry out \textbf{in-depth analyses of performance differences and potential mechanisms}, which sheds light on the future development of multimodal intelligence towards AGI.

\section{Related Work}
\subsection{Machine Unlearning for LLMs}
Initially developed for classification tasks, MU for LLMs has recently gained attention as a response to concerns regarding the unintended memorization of pretraining data~\cite{si2023llmu_survey}. The majority rely on \textbf{parameter optimization-based methods}~\cite{nguyen2022mu_survey}, such as Gradient Ascent~\cite{thudi2022GA} and its variations~\cite{liu2022GA_Diff}. While fine-tuning via cross-entropy loss remains a common practice, specific loss functions like KL minimization~\cite{nguyen2020KL_Min,wang2023kga,liu2024revisiting} and IDK~\cite{maini2024tofu} have been designed to better control the outputs of unlearned models. Besides,~\citet{zhang2024npo} reframe LLM unlearning as a preference optimization problem~\cite{rafailov2024dpo}, applying Negative Preference Optimization loss to enhance the unlearning. \par 

In addition, MU algorithms that \textbf{do not alter internal parameters} have also been explored. These include approaches based on model editing~\cite{ilharco2022editing,wu2023depn}, task vectors~\cite{eldan2023whp,li2024wmdp}, or in-context learning~\cite{pawelczyk2023icun,thaker2024guardrail}. While free from tuning, they often fail to achieve a sufficient level of unlearning or incur higher computational costs for detecting privacy units~\cite{ilharco2022editing,wu2023depn}.

\subsection{Multimodal Machine Unlearning}\label{MMU}
Before the development of MLLMs, research on MU in multimodal models primarily focused on Vision-Language Models ~\cite{radford2021clip} and Text-to-Image models~\cite{rombach2022ldm}. For encoder-decoder models~\cite{li2021albef,li2022blip}, MultiDelete~\cite{cheng2024multidelete} introduces a method that separates cross-modal embeddings for the forget set while preserving unimodal embeddings for the retain set. Additionally, ~\citet{yang2024cliperase} achieves class-wise forgetting in CLIP by fine-tuning selected salient layers solely on synthetic samples. In the context of T2I models, several pioneering studies~\cite{Gandikota2023ErasingCF,Zhang2023ForgetMeNotLT} have discussed to delete specific concepts, such as not-safe-for-work (NSFW) content, within diffusion models. Among these, SalUn~\cite{Fan2023SalUnEM} and SFR-on~\cite{Huang2024UnifiedGM} selectively update salient parameters to balance the dual objectives of maintaining generalization and ensuring efficient data forgetting. 

Despite these advancements, MU for MLLMs remains in its nascent stages. Specifically, SIU~\cite{li2024siu} investigates the erasure of visual patterns in MLLMs using the real-world entity dataset MMUBench through multifaceted fine-tuning. There are also discussions about the application of MU in MLLMs, including hallucination mitigation~\cite{Xing2024EFUFEF} and safety alignment~\cite{Chakraborty2024CrossModalSA}.

See more related work in Appendix \ref{app:rk}.

\section{Our Proposed \method}
\subsection{Task Setting}\label{sec:task}
\begin{figure*}[thp]
\centering
\includegraphics[width=0.98\textwidth]{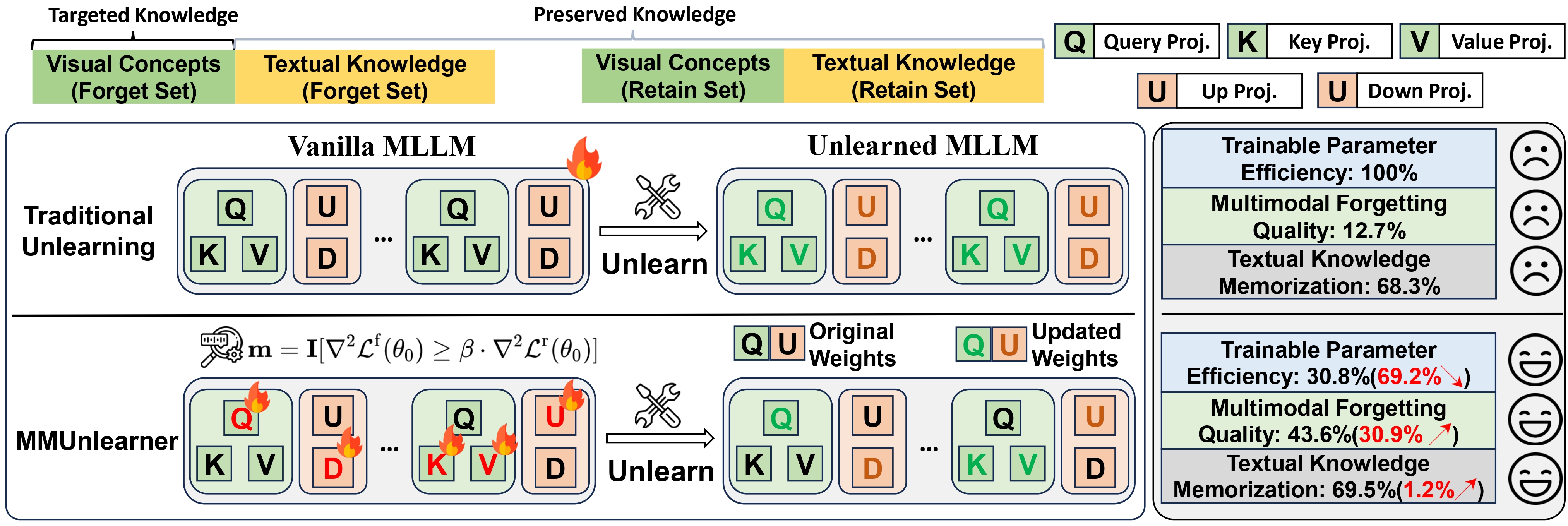}
\caption{An illustration of our proposed \method. Compared to traditional approaches employed in previous work, which directly apply LLM-based unlearning algorithms to vanilla MLLMs, our method demonstrates superior parameter efficiency, forgetting performance, and textual knowledge preservation. Both the baseline and our approach are trained on VQA-format data, while textual QA-format data is used to assess the preservation of textual knowledge during evaluation.}
\label{fig:method}
\end{figure*}
% A image contains hierarchical concepts from general to fine-grained. For example, "Trump" is composed by "blone hair", "red tie" and others. Those concepts are connected with the textual knowledge within LLMs (https://arxiv.org/abs/2412.14133). We need to erase targeted high concept C_h while preserve common concepts.
To enable a text-only LLM $\mathcal{L}$ to comprehend visual context, mainstream approaches extract visual embeddings $H_I$ using a vision encoder $\mathcal{V}$ followed by a projector $\mathcal{W}$. The entire model is then fine-tuned with visual instruction data $\{X_I, X_Q, X_A\}$, where $X_I$ represents the input image, $X_Q$ is the textual instruction, and $X_A$ denotes the expected answer of length $S$. This can be formalized as follows:
\begin{equation}\label{eq:train}
\begin{split}
    X_O &= \mathcal{L}(H_I; X_Q) = \mathcal{L}(\mathcal{W}(\mathcal{V}(X_I)); X_Q),\\
    {Loss} &= - \sum_{s=1}^{S} \log P(X_A^{(s)} | X_O^{(<s)}),
\end{split}
\end{equation}
where $X_O$ represents the output sequence of the model, $X_A^{(s)}$ is the target token at step $t$, and $X_O^{(<s)}$ represents the previously generated tokens in the output sequence. $P(X_A^{(s)} | X_O^{(<s)})$ represents the predicted probability of label $X_A^{(s)}$ at position $s$. Following this framework, MLLMs acquire the ability to recognize concepts in the visual modality, establishing associations between visual concepts and the internal knowledge of the LLM while leveraging its reasoning and generative capabilities.

\begin{tcolorbox}[float=t!,title=Desiderata for MLLM-based Multimodal Machine Unlearning]\label{box:desiderata}
    \begin{itemize}
        \item[\textbf{I:}] \textbf{Forgetting $C$ in the visual modality.} The model should fail to recognize concept $C$ in visual inputs, i.e., $\mathcal{L}(h_I; x_{Q_v,C}) \neq x_{A_v,C}$, where $x_{Q_v,C}$ is a textual query referring to $C$ in the image, and $x_{A_v,C}$ is the correct answer to $x_{Q_v,C}$.
        
        \item[\textbf{II:}] \textbf{Preserving general visual perception abilities.} The model should retain its ability to process visual information unrelated to concept $C$, i.e., $\mathcal{L}(h_I; x_{Q_v,\sim C}) = x_{A_v,\sim C}$, where $x_{Q_v,\sim C}$ is a textual query unrelated to $C$ in the image, and $x_{A_v,\sim C}$ is the correct answer to $x_{Q_v,\sim C}$.
        
        \item[\textbf{III:}] \textbf{Retaining internal knowledge within the LLM.} The model should preserve its textual knowledge about concept $C$, i.e., $\mathcal{L}(x_{Q_t,C}) = x_{A_t,C}$, where $x_{Q_t,C}$ is a pure-text query about concept $C$, and $x_{A_t,C}$ is the correct answer to $x_{Q_t,C}$.
    \end{itemize}
\end{tcolorbox}

The objective of multimodal MU is, therefore, to eliminate these learned associations between a specific concept and its corresponding visual patterns. In other words, the unlearned model should behave as if it has never encountered the related images during the visual instruction tuning process. Specifically, for a given concept $C$, its image representation $x_I$, and the extracted visual embeddings $h_I = \mathcal{W}(\mathcal{V}(x_I))$, the unlearned model must satisfy the conditions in 
Box \ref{box:desiderata}.

In standard unlearning tasks, the unlearned model is expected to maintain its knowledge on a retained set, which we denotes in next Section \ref{sec:floss}. We illustrate the expected behavior of the targeted model using biographical examples, though our task formulation can be extended to other domains, such as real-world entities and landmarks.

\subsection{Selective Updating for Forget Loss}\label{sec:floss}
The core challenge of multimodal MU lies in preserving textual knowledge while performing unlearning on VQA data. A naive unlearning method, such as GA Difference (\textbf{GA\_Diff}), simply updates the model parameters $\theta$ using a joint loss ($\mathcal{L}^J(\cdot)$) computed over the Forget VQA set $D_f=\{(I_{C_f}, X_{Q_v}, X_{A_v}, C_f)\}$ and the Retain VQA set $D_r=\{(I_{C_r}, X_{Q_v}, X_{A_v}, C_r)\}$ as follows:
\begin{equation}
\begin{split}
\mathcal{L}^J(\theta_t) = -\mathcal{L}^{f}(\theta_t) + \mathcal{L}^{r}(\theta_t),
\end{split}
\end{equation}
where $t$ denotes the $t$-th step, and $\mathcal{L}^{f}(\cdot)$ and $\mathcal{L}^{r}(\cdot)$ represent the loss on the Forget and Retain sets, respectively. Interpreting the update of $\theta$ as an optimization problem in parameter space, the term $-\mathcal{L}^{f}(\theta_t)$ forces the MLLM to forget the VQA samples that should be unlearned by following the steepest ascent direction. And $\mathcal{L}^{r}(\theta_t)$ aims to preserve knowledge from the retained VQA samples. However, the conflicting directions of the Forget loss and the Retain loss make the unlearning process unstable. Furthermore, traditional MLLM unlearning methods primarily focus on VQA data, neglecting the constraints from text-only QA data.\par

Nonetheless, such conflicts can be effectively mitigated if model updates selectively target parameters that are salient for the targeted knowledge (\textbf{S}) while preserving those critical for others. This process can be formulated as:
\begin{equation}
\begin{split}
\mathcal{L}^S(\theta_t) = -\mathbf{m} \odot \mathcal{L}^{f}(\theta_t) + \mathcal{L}^{r}(\theta_t),
\end{split}
\end{equation}
where $\mathbf{m}$ is a boolean mask that selectively updates parameters, and $\odot$ denotes the Hadamard product. In this way, the ascent of the Forget Loss on targeted visual concept does not destroy the parameters salient for the Retain set or textual knowledge, as illustrated in Figure \ref{fig:method}.

\subsection{Weight Saliency Map in Parameter Space}
As discussed in Section \ref{sec:floss}, the gradient mask $\mathbf{m}$ should strike a balance between forgetting and retaining knowledge so that only the necessary parameters are updated during unlearning. Inspired by \citet{fan2023salun,huang2024remain}, the saliency map of each parameter on a given dataset $D$ in the parameter space can be approximated by the diagonal of the initial model’s Fisher information matrix:
\begin{equation}
\begin{split}
S(\theta_0, \mathcal{L}, D) = F_{diag}^D = [\nabla\mathcal{L}^D(\theta_0)]^2,
\end{split}
\end{equation}
which corresponds to a manifold defined by the loss function, dataset distribution, and initial parameters in the parameter space.

From this perspective, we define a targeted dataset as:
\begin{equation}
T = \{(I_{C_f}, X_{Q_v}, X_{A_v}, C_f)\},
\end{equation}
while the preserved dataset is defined as:
\begin{equation}\label{eq:pdata}
\begin{split}
P &= \{(X_{Q_t}, X_{A_t}, C_f)\}\cup \{(X_{Q_v}, X_{A_v}, C_r)\}\\ & \cup \{(I_{C_r}, X_{Q_t}, X_{A_t}, C_r)\},
\end{split}
\end{equation}
where $C_f$ represents the targeted concepts to be forgotten, and $C_r$ denotes the untargeted concepts that should be retained. Here, $I$, $X_{Q_{v/t}}$, and $X_{A_{v/t}}$ represent the corresponding image, multimodal/pure-textual query, and correct answers, respectively.

Thus, the gradient mask $\mathbf{m}$ is obtained by comparing the relative ratio of the saliency map between the targeted and preserved datasets using a hard threshold:
\begin{equation}\label{eq:mask}
\begin{split}
\mathbf{m} &= \mathds{1}\left[\frac{S(\theta_0, \mathcal{L}, T)}{S(\theta_0, \mathcal{L}, P)} \geq \beta \right]\\
&= \mathds{1}\left[\frac{\nabla^2\mathcal{L}^T(\theta_0)}{\nabla^2\mathcal{L}^P(\theta_0)} \geq \beta \right],
\end{split}
\end{equation}
where $\mathds{1}[y \geq \beta]$ is an element-wise indicator function that outputs 1 for the $i$-th element if $y_i \geq \beta$ and 0 otherwise. The threshold $\beta > 0$ is a hard cutoff; for simplicity, we use $\beta=1$ throughout our experiments, which is sufficient for our tasks.

\section{Experiment}
\subsection{Experiment Settings}
\begin{table*}[t!]
    \centering
\scalebox{0.98}{
\begin{tabular}{l|cccccc|cccccc}
\toprule
\multirow{3}{*}{\textbf{Methods}} 
& \multicolumn{6}{c|}{\textbf{MLLMU-Bench}} 
& \multicolumn{6}{c}{\textbf{CLEAR}} \\
\cline{2-13}
    & \begin{tabular}[c]{@{}c@{}}Forget VQA.\\ Acc (\textcolor{blue}{$\downarrow$})\end{tabular}
     & \begin{tabular}[c]{@{}c@{}}Forget QA.\\ Acc (\textcolor{red}{$\uparrow$})\end{tabular}
     & \begin{tabular}[c]{@{}c@{}}Retain VQA.\\ Acc (\textcolor{red}{$\uparrow$})\end{tabular}
     & \begin{tabular}[c]{@{}c@{}}Retain QA.\\ Acc (\textcolor{red}{$\uparrow$})\end{tabular}
     & \begin{tabular}[c]{@{}c@{}}Realworld VQA.\\ Acc (\textcolor{red}{$\uparrow$})\end{tabular}
     & \begin{tabular}[c]{@{}c@{}}Realworld QA.\\ Acc (\textcolor{red}{$\uparrow$})\end{tabular}
     
     & \begin{tabular}[c]{@{}c@{}}Forget VQA.\\ Acc (\textcolor{blue}{$\downarrow$})\end{tabular}
    & \begin{tabular}[c]{@{}c@{}}Forget QA.\\ ROUGE-L (\textcolor{red}{$\uparrow$})\end{tabular}
    & \begin{tabular}[c]{@{}c@{}}Retain VQA.\\ Acc (\textcolor{red}{$\uparrow$})\end{tabular}
    & \begin{tabular}[c]{@{}c@{}}Retain QA.\\ ROUGE-L (\textcolor{red}{$\uparrow$})\end{tabular}
    & \begin{tabular}[c]{@{}c@{}}Realworld VQA.\\ Acc (\textcolor{red}{$\uparrow$})\end{tabular}
    & \begin{tabular}[c]{@{}c@{}}Realface VQA.\\ Acc (\textcolor{red}{$\uparrow$})\end{tabular}\\
\midrule
\multicolumn{13}{c}{\textbf{LLaVA-1.5-7B}} \\
\midrule
Vanilla&45.8\%&38.4\%&45.2\%&37.5\%&47.4\%&54.9\%&63.3\%&0.367&54.0\%&0.352&53.7\%&85.4\%\\
\hdashline
GA&43.2\%&32.5\%&\underline{45.0\%}&\underline{32.2\%}&\underline{47.0\%}&\textbf{55.0\%}&57.4\%&0.153&\textbf{52.4\%}&0.176&51.8\%&\underline{83.4\%}\\
GA\_Diff&\underline{40.0\%}&\underline{33.6\%}&44.3\%&31.5\%&46.6\%&53.6\%&47.3\%&0.197&43.4\%&0.220&47.7\%&73.5\%\\
KL\_Min&42.4\%&\underline{33.6\%}&44.9\%&32.0\%&\textbf{47.4\%}&54.6\%&\underline{40.4\%}&0.270&38.1\%&0.274&51.5\%&82.8\%\\
NPO&43.2\%&\underline{33.6\%}&\textbf{45.2\%}&\underline{32.2\%}&\underline{47.0\%}&55.0\%&\underline{40.4\%}&\underline{0.285}&38.6\%&\underline{0.282}&\textbf{52.9\%}&\underline{83.4\%}\\
\rowcolor{black!20}Ours&\textbf{31.2\%}&\textbf{34.2\%}&44.2\%&\textbf{35.1\%}&46.7\%&\underline{54.9\%}&\textbf{36.2\%}&\textbf{0.348}&\underline{46.6\%}&\textbf{0.338}&\underline{52.3\%}&\textbf{84.1\%}\\
\midrule

\multicolumn{13}{c}{\textbf{Qwen2-VL-7B}} \\
\midrule
Vanilla&55.2\%&55.0\%&56.0\%&58.6\%&77.3\%&77.5\%&67.0\%&0.116&70.9\%&0.098&69.2\%&91.4\%\\
\hdashline
GA&50.4\%&46.7\%&\underline{51.5\%}&\textbf{57.6\%}&74.4\%&\underline{77.8\%}&\underline{55.3\%}&\underline{0.123}&62.4\%&0.083&65.9\%&86.8\%\\
GA\_Diff&54.4\%&\underline{52.8\%}&38.8\%&54.4\%&74.5\%&77.0\%&63.3\%&\textbf{0.125}&\textbf{71.4\%}&0.088&\textbf{70.0\%}&\underline{92.7\%}\\
KL\_Min&\underline{45.6\%}&45.3\%&35.9\%&55.6\%&74.8\%&77.1\%&67.0\%&0.120&\underline{70.9\%}&\underline{0.098}&68.4\%&90.7\%\\
NPO&49.6\%&50.4\%&49.5\%&53.3\%&\underline{75.2\%}&\textbf{78.3\%}&62.8\%&0.103&68.3\%&0.091&\underline{68.9\%}&88.7\%\\
\rowcolor{black!20}Ours&\textbf{44.0\%}&\textbf{54.4\%}&\textbf{56.0\%}&\underline{55.7\%}&\textbf{75.3\%}&77.3\%&\textbf{50.0\%}&\underline{0.123}&\underline{70.9\%}&\textbf{0.100}&\underline{68.9\%}&\textbf{94.7\%}\\
\bottomrule
\end{tabular}}
    \vspace{-0.1in}
    \caption{Overall results of baselines and \method on two representative MLLMs across two unlearning benchmarks. \textbf{Bold} indicates the best performance, and \underline{underline} denotes the runner-up. Each baseline method is evaluated on six dimensions among each dataset, assessed by classification accuracy (\textit{i.e.,} Acc) for multi-choice QA task and ROUGE-L score for generation task. \textcolor{blue}{$\downarrow$} indicates that lower values are better, while \textcolor{red}{$\uparrow$} indicates that higher values are better. More results can be found in Appendix \ref{app:additional}.}
    \label{tab:main-table}
\end{table*}

\subsubsection{Datasets and Metrics}
To demonstrate the effectiveness of our proposed \method, we conduct experiments on two MLLM-based unlearning benchmarks:

\textbf{MLLMU-Bench}~\cite{liu2024mllmubench}. It consists of fictitious personal profiles, each accompanied by a portrait and 14 corresponding questions (\textit{i.e.,} 7 VQA questions and 7 textual QA questions) with multiple-choice options. For the Forget, Retain, and Real-world sets used in our experiments, we report the \textit{average accuracy} as the metric.

\textbf{CLEAR}~\cite{dontsov2024clear}. It is built on top of TOFU~\cite{maini2024tofu}, a dataset containing fictional author profiles designed for LLM unlearning. For each author in TOFU, CLEAR adds several face images to it, along with captions generated by GPT-4o~\cite{OpenAI2023GPT4TR}. In our experiments, we evaluate the Forget, Retain, and Real-world sets using \textit{average accuracy} for VQA task and \textit{ROUGE-L}~\cite{Lin2004ROUGEAP} for textual QA task, respectively. Note that \textbf{only} VQA data is used for unlearning tuning in both datasets, while textual QA data is used solely for evaluation across different baselines, aligning with previous works. Please refer to Appendix \ref{app:dataset} and \ref{app:metrics} for details of the datasets and evaluation metrics.\par

\subsubsection{Evaluated MLLMs}
To further verify the generalizability of our conclusions, we use two MLLMs, LLaVA-1.5-7B-hf\footnote{\url{https://huggingface.co/llava-hf/llava-1.5-7b-hf}} and Qwen2-VL-7B-Instruct\footnote{\url{https://huggingface.co/Qwen/Qwen2-VL-7B-Instruct}}, as our base models. The vanilla models used for unlearning are trained following the official implementations provided by MLLMU-Bench\footnote{\url{https://github.com/franciscoliu/MLLMU-Bench}} and CLEAR\footnote{\url{https://github.com/somvy/multimodal_unlearning}} respectively. More details can be found in Appendix \ref{app:base_hyper} and \ref{app:vanilla}.
% To further verify the generalizability of our conclusions, we use two MLLMs, LLaVA-1.5-7B-hf and Qwen2-VL-7B-Instruct, as our base models. The vanilla models used for unlearning are trained following the official implementations provided by MLLMU-Bench and CLEAR respectively. More details can be found in Appendix \ref{app:base_hyper} and \ref{app:vanilla}.

\subsubsection{Baselines}
Following~\citet{liu2024mllmubench}, we compare our method with the following four baselines: 

\textbf{GA}~\cite{thudi2022GA} applies opposite gradient updates on Forget VQA set $D_f$. 

\textbf{GA\_Diff}~\cite{liu2022GA_Diff}, an improved variant of GA, introduces joint loss to make a balance between $D_f$ and Retain VQA set $D_r$, as discussed in Section \ref{sec:floss}. 

\textbf{KL\_Min}~\cite{maini2024tofu} aligns the model’s predictions on $D_r$ with those of the original model while encouraging divergence from the Forget Set, implementing by minimizing the KL Divergence. 

\textbf{NPO}~\cite{zhang2024npo} treats $D_f$ as dispreferred data and casts unlearning into a preference optimization framework, with an oracle model fine-tuned exclusively on $D_r$. 

Our implementations are based on the official code from MLLMU-Bench and CLEAR, with the same pipeline. Considering that visual concepts can be stored in the vision encoder in real-world cases, we carry out our experiments with parameters of both vision encoder and language model trainable. Details of baselines can be found in Appendix \ref{app:formula}.

\subsubsection{Implementation Details}
All the experiments including fine-tuning and baseline implementation of LLaVA 1.5 and Qwen2-VL were conducted on the A800 GPU cluster, with full precision used. For a fair comparison, we set the same learning rate, unlearning epochs, and batch size across all methods (details in Appendix \ref{app:base_hyper}).
\subsection{Main Result}
In this section, we present the performance of MU methods on MLLMU-Bench and CLEAR dataset, offering a comprehensive comparison between four baselines and \method, as detailed in Table \ref{tab:main-table}. To validate the generalizability and efficiency of \method, we further analyze the relationship between forget ratios and various metrics. Overall, our method provides a more accurate yet efficient approach to erasing visual concepts. The key observations are as follows:

% Note that tuning on the CLEAR dataset is quite unstable, and unlearned models can easily crash completely (yielding a 0 ROUGE-L score for generation tasks), as noted in the original paper~\cite{dontsov2024clear}. Therefore, to ensure our conclusions are both persuasive and reliable, we report the best performance of the baselines in our experiments on the CLEAR dataset.
% In this section, we present a comprehensive comparison of various unlearning algorithms across different forget data splits using the MLLMU-Bench benchmark 

\ding{182} \textbf{\method excels in erasing visual concepts.}
For MLLMU-Bench, our method achieves the lowest accuracy on the Forget VQA Set for both LLaVA-7B and Qwen2-VL, demonstrating the efficiency of \method. Compared to the Vanilla model, \method shows a significant accuracy drop of 14.6\% and 11.2\%, respectively, outperforming all baseline methods. For CLEAR, our method also improves accuracy on the Forget VQA Set by 4.2\% and 5.3\% compared to the best baseline results. This highlights the effectiveness of \method in erasing targeted visual concepts.

\ding{183} \textbf{\method preserves untargeted visual concepts from Retain VQA and overall textual knowledge effectively.}
Despite its superior unlearning capability, \method also demonstrates outstanding performance in preserving untargeted knowledge. Specifically, it achieves state-of-the-art results on the Retain Set and Forget QA Set in most cases, particularly for LLaVA-7B on MLLMU-Bench QA and CLEAR QA. In other tasks, such as Retain VQA and real-world VQA, \method remains highly competitive, with performance gaps of no more than 2\% from the best baseline results, except for a 5.8\% drop behind GA on the Retain QA of CLEAR. However, considering the poor Forget VQA performance of GA on CLEAR compared to other baselines, we consider this deviation reasonable.

\ding{184} \textbf{Existing baselines struggle with unlearning visual concepts, although relatively better on textual knowledge removal.}\label{con:text}
We find that most baseline methods effectively remove textual knowledge but struggle to erase learned visual concepts. For example, NPO achieves the best trade-off between Forget VQA and Retain VQA, performing the best on the Forget VQA Set while maintaining strong performance on the Retain VQA Set. However, even NPO shows a bias toward textual QA data, as its accuracy drop on the Forget QA Set is significantly larger than that on the Forget VQA Set for MLLMU-Bench. \textbf{The success of baselines in textual knowledge removal aligns with previous findings~\cite{liu2024mllmubench}, yet their inefficacy in handling visual concepts underscores the need for dedicated MU algorithms tailored for MLLMs}, rather than merely adapting LLM-oriented MU methods to VQA data.

\begin{figure*}[!t]
\centering
\vspace{-0.1in}
\includegraphics[width=\textwidth]{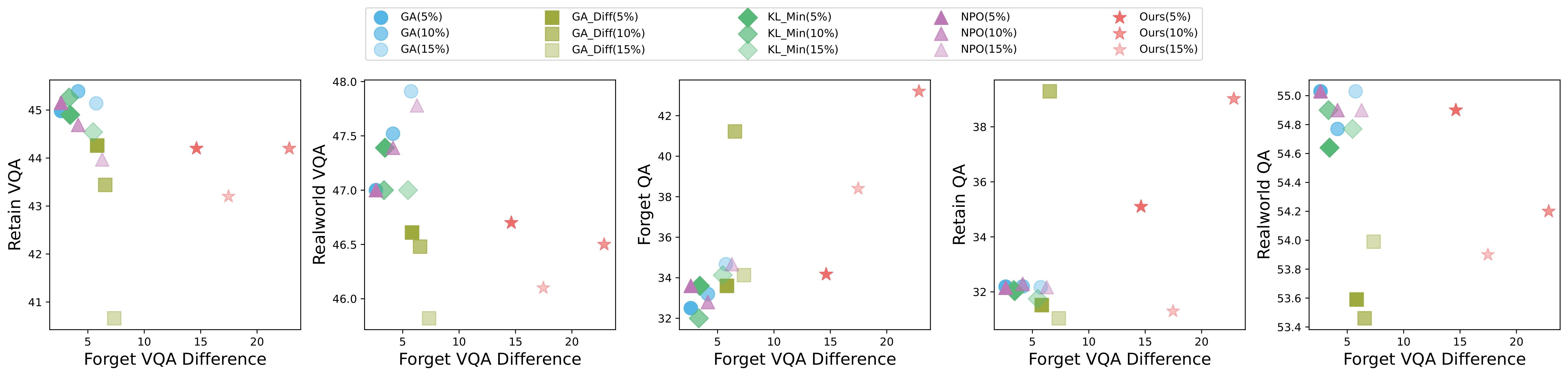}
\vspace{-0.25in}
\caption{
The overall trade-off between unlearning effectiveness and model utility across five dimensions under varying forget ratios, using LLaVA as the base model. The $x$-axis represents the change in forget classification accuracy relative to the vanilla model, while the $y$-axis captures model utility from multiple perspectives. From left to right, these perspectives encompass Retain VQA, Real-world VQA, Forget QA, Retain QA, and Real-world QA performance.}
\vspace{-0.1in}
\label{fig:trade-off}
\end{figure*}

\subsection{Unlearning v.s. Model Utility}
Previous works on LLM unlearning~\cite{zhang2024npo,liu2024towards} and MLLM unlearning~\cite{liu2024mllmubench} have discussed the trade-off between unlearning effectiveness and model utility as the forget ratio varies. However, textual utility in MLLM unlearning remains largely unexplored. In this section, we analyze the performance of different methods across three forget ratios (\textit{i.e.,} 5\%, 10\%, and 15\%), as shown in Figure \ref{fig:trade-off}.

\ding{182} \textbf{\method remains efficient across different forget ratios.} \method demonstrates remarkable forgetting performance across various forget ratios. In most cases, the difference in Forget VQA accuracy between \method and the vanilla model surpasses other baselines by a significant margin, ranging from 5\% to 15\%. Among the four baselines, GA\_Diff exhibits the strongest capability in erasing visual concepts, while NPO achieves competitive results at higher forget ratios. Notably, as the forget ratio increases, all baselines show improvements in forget quality, albeit at the cost of degraded model utility on Retain and Real-world tasks. Furthermore, the trend of \method in relation to the forget ratio presents similar pattern with that of GA\_Diff, but with superior forget quality and lower utility decay, as reflected in Retain VQA, Real-world VQA, Forget QA, and Retain QA.\par

\ding{183} \textbf{Higher forget ratio makes it harder to maintain Model Utility.} There is a clear downward trend in model utility for VQA tasks as the forget ratio increases. When the forget ratio rises from 5\% to 15\%, GA\_Diff experiences the most significant drop, with over a 3\% decrease in Retain VQA performance compared to other baselines. However, by selectively updating the vanilla model using a weight saliency map, \method effectively mitigates this issue, achieving a better trade-off between forgetting and retention. A similar phenomenon can be observed for KL\_min, NPO, and GA. Additionally, performance on Real-world VQA exhibits the smallest variation across all methods, indicating the robustness of the visual features learned by MLLMs.\par

\begin{table}[t!]
    \centering
\scalebox{0.98}{
\begin{tabular}{l|cc|cc|cc}
\toprule
\multirow{3}{*}{\textbf{Modules}} 
& \multicolumn{2}{c|}{\textbf{Forget Set}} 
& \multicolumn{2}{c|}{\textbf{Retain Set}}
& \multicolumn{2}{c}{\textbf{Realworld Set}} \\
\cline{2-7}
    & \begin{tabular}[c]{@{}c@{}}Forget VQA.\\ Acc (\textcolor{blue}{$\downarrow$})\end{tabular}
     & \begin{tabular}[c]{@{}c@{}}Forget QA.\\ Acc (\textcolor{red}{$\uparrow$})\end{tabular}
     & \begin{tabular}[c]{@{}c@{}}Retain VQA.\\ Acc (\textcolor{red}{$\uparrow$})\end{tabular}
     & \begin{tabular}[c]{@{}c@{}}Retain QA.\\ Acc (\textcolor{red}{$\uparrow$})\end{tabular}
     & \begin{tabular}[c]{@{}c@{}}Realworld VQA.\\ Acc (\textcolor{red}{$\uparrow$})\end{tabular}
     & \begin{tabular}[c]{@{}c@{}}Realworld QA.\\ Acc (\textcolor{red}{$\uparrow$})\end{tabular}\\
\midrule
Vanilla&45.8\%&38.4\%&45.2\%&37.5\%&47.4\%&54.9\%\\
\hdashline
LM+Connector&30.4\%&33.4\%&43.2\% &36.9\%& 46.5\%& 53.9\% \\
Vision Encoder &33.6\%& 33.0\%&  42.4\%& 37.5\%& 38.3\%& 51.1\% \\
All & 31.2\%&  34.2\%&  44.2\%&   35.1\%&  46.7\%&  54.9\% \\
\bottomrule
\end{tabular}}
    \vspace{-0.1in}
    \caption{Results for updating different modules of MLLMs with \method. We abbreviate the language model as LM. The vision encoder has been updated during obtaining Vanilla model to simulate real-world settings.}
    \label{tab:ablation}
    \vspace{-4mm}
\end{table}

\ding{184} \textbf{\method show powerful ability on textual knowledge preservation even under high forget ratio.} The scatter plots of the four baselines cluster in the lower-left region on Forget QA and Retain QA, aligning with the findings in Observation \ref{con:text}. In contrast, \method exhibits a remarkable ability to preserve textual knowledge from both the Forget and Retain sets, significantly outperforming the baselines. Surprisingly, \method's performance on QA tasks improves as the forget ratio increases. We attribute this phenomenon to the enhanced influence of the preserved dataset $P$ in Eq. \ref{eq:pdata}, which becomes more effective as more forget samples are considered.

\begin{table*}[!t]
  \centering
  % \SetTblrInner{rowsep=0.1pt}
  \begin{tblr}{
    colspec = {p{0.33\linewidth} p{0.33\linewidth} p{0.33\linewidth}},
  }
    \toprule
    {\textbf{Visual Concept Removal}}
      & {\textbf{Textual Knowledge Preservation}}
      & {\textbf{Untargeted Concept Retention}}\\
    \midrule
      \small{
      Info: \textbf{ID=144, Forget VQA}\\
      Q: \textcolor{mygreen}{<image>}What profession does the individual in the image pursue?\\
      A: The individual in the image is \textcolor{myblue}{an architect.}\\
      GA: This person is \textcolor{myblue}{an architect.}\\
      GA\_Diff: This person is \textcolor{myblue}{an architect.}\\
      KL\_Min: This person is \textcolor{myblue}{an architect.}\\
      NPO: This person is \textcolor{myblue}{an architect.}\\
      Ours: This person is \textcolor{myred}{a marine biologist.}}
      & \small{
      Info: \textbf{ID=144, Forget QA}\\
      Q: What year was Anika Graves born?\\
      A: Anika Graves was born in \textcolor{myblue}{1985.}\\
      GA: Anika Graves was born in \textcolor{myred}{1992.}\\
      GA\_Diff: Anika Graves was born in \textcolor{myred}{1992.}\\
      KL\_Min: Anika Graves was born in \textcolor{myred}{1992.}\\
      NPO: Anika Graves was born in \textcolor{myred}{1992.}\\
      Ours: Anika Graves was born in \textcolor{myblue}{1985.}}
      & \small{
      Info: \textbf{ID=437, Retain VQA}\\
      Q: \textcolor{mygreen}{<image>}What profession does this person appear to be associated with?\\
      A: The person is likely \textcolor{myblue}{an architect.}\\
      GA: This person is associated with \textcolor{myred}{the field of environmental science.}\\
      GA\_Diff: This person is associated with \textcolor{brown}{the field of architecture.}\\
      KL\_Min: This person is associated with \textcolor{myred}{the field of environmental science.}\\
      NPO: This person is associated with \textcolor{myred}{the field of environmental science.}\\
      Ours: This person is \textcolor{myblue}{an architect.}}\\
    \bottomrule
  \end{tblr}
  \caption{Illustration of some of the most challenging visual concepts to forget. \textcolor{myblue}{$\sbullet[.75]$} and \textcolor{myred}{$\sbullet[.75]$} indicate correct and incorrect answers, respectively. \textcolor{brown}{$\sbullet[.75]$} denotes paraphrased answer while \textcolor{mygreen}{$\sbullet[.75]$} highlights image inputs.}
  \label{tab:cases}
  \vspace{-2mm}
\end{table*}

\subsection{Ablation Study}

Considering real-world scenarios where visual concepts can be learned by the vision encoder through pre-training and supervised fine-tuning~\cite{goh2021clipneuron}, we keep the vision encoder's parameters trainable both when obtaining the vanilla model and during the unlearning process, following previous practices~\cite{wang2024qwen2vl,lu2024deepseek}. To analyze the impact of unlearning on different modules, we conduct an ablation study on LLaVA-7B using MLLMU-Bench, with the results summarized in Table \ref{tab:ablation}. While there are minor differences in performance depending on which modules are updated during unlearning, we argue that all configurations achieve competitive results. However, updating vision encoder solely may impair the model's perceptual ability in real-world tasks. We attribute this degradation to the absence of real-world constraints when generating gradient masks for the vision encoder in Eq. \ref{eq:mask}.
\subsection{Case Study}

In this section, we illustrate the performance of \method on a given visual concepts, and compared it with GA and NPO. As shown in Table \ref{tab:cases}, \method exceeds other baselines in targeted visual concept removal, textual knowledge preservation and untargeted concept retention. More detailed cases can be found in Appendix \ref{app:case}.

\subsection{Visualization}
\begin{figure}[!t]
\centering
\includegraphics[width=\linewidth]{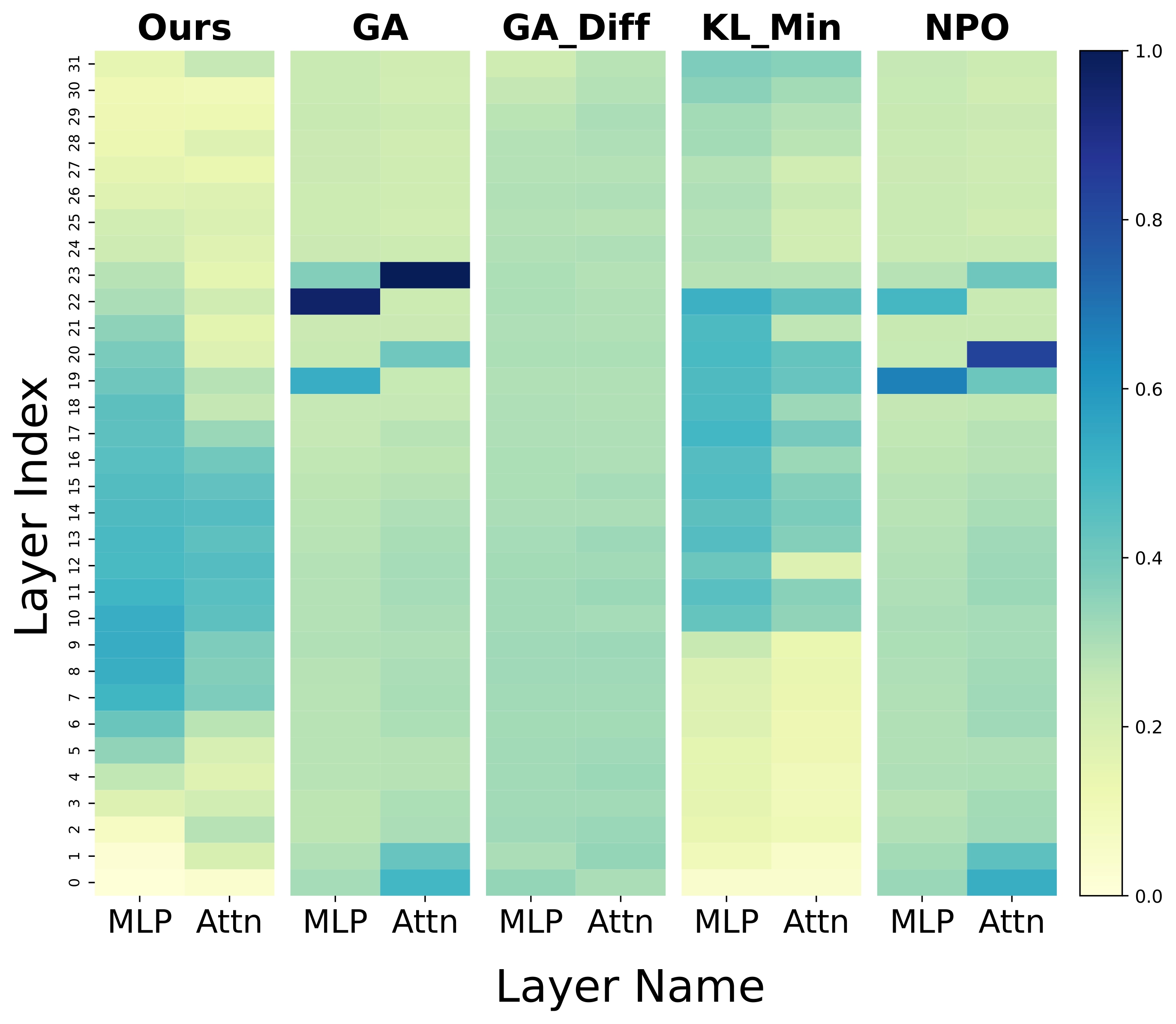}
\vspace{-0.1in}
\caption{ 
The distribution of the top-$n$ deviated parameters across different MU algorithms for LLaVA, where $n$ corresponds to the number of unmasked parameters in Eq. \ref{eq:mask}. The $x$-axis represents different model layers while the $y$-axis denotes the layer index. Color reflects density of updated parameters, with darker colors for higher percentage of updates.}
\vspace{-0.1in}
\label{fig:heatmap}
\end{figure}
We visualize the parameter distribution selected by \method through a heatmap, comparing it against other unlearning methods by selecting the top-$n$ parameters with the largest deviation post-unlearning. As shown in Figure \ref{fig:heatmap}, which presents results on LLaVA-7B using MLLMU-Bench, GA and NPO exhibit similar update patterns, primarily affecting middle MLP, middle Attention, and shallow Attention layers. In contrast, \method produces a more focused and structured distribution, peaking in the middle MLP and Attention layers. According to prior MLLM interpretability studies~\cite{basu2024understanding,yu2024understanding}, shallow Attention layers are crucial for visual information transfer, while the middle MLP layers handle information storage and aggregation. Our findings align well with previous research, providing possible insights into the distinctions among different unlearning methods for MLLMs. However, a more in-depth exploration of unlearning mechanisms is left for future work. Additional visualizations across different models and datasets are provided in the Appendix \ref{app:viz}.

\section{Conclusion}
In this paper, we reformulate the task of MU tailored for MLLMs, a field still in its early stages. Our proposed setting aims to erase targeted visual concepts in MLLMs while preserving untargeted knowledge. To address this challenge, we further propose a novel weight saliency-based unlearning method, \method, which selectively updates parameters crucial for the forgetting objective while protecting parameters essential for retaining untargeted knowledge. Our experiments demonstrate that directly transferring LLM-oriented MU methods to VQA data is insufficient for MLLMs; whereas our proposed \method exhibits a strong ability to remove visual concepts while preserving textual knowledge. Further experiments validate the effectiveness and robustness of our approach. We believe that \method will lay a solid foundation for building a trustworthy MLLM ecosystem to achieve ultimate AGI.

\section{Acknowledgements}
This work was supported by Guangdong Provincial Department of Education Project (Grant No.2024KQNCX028); Scientific Research Projects for the Higher-educational Institutions (Grant No.2024312096), Education Bureau of Guangzhou Municipality; Guangzhou-HKUST(GZ) Joint Funding Program (Grant No.2025A03J3957), Education Bureau of Guangzhou Municipality.
% \newpage
% \clearpage
\section{Limitations}
% 1. Dataset, leave for future work.
% 2. There is still some degradation in utility compared to the original model.
% 3. Mechanism explanation.
Despite the contributions demonstrated in our work, several limitations remain:
\begin{itemize}
\item [1.] While we provide a detailed analysis of various unlearning methods, our experiments primarily focus on MLLMU-Bench~\cite{liu2024mllmubench} and CLEAR~\cite{dontsov2024clear}, two pioneering benchmarks for MLLM MU. As this field is still in its early stages, designing more high-quality benchmarks would be beneficial for evaluating MLLM-targeted unlearning methods more comprehensively. For instance, representative LLM unlearning benchmarks such as TOFU~\cite{maini2024tofu} and WPU~\cite{liu2024revisiting} could be extended with visual information, facilitating a more thorough assessment of MLLM MU. However, we leave the enhancement and development of MLLM-oriented unlearning benchmarks for future work.\par
\item [2.] Although \method surpasses baseline methods in forgetting tasks, there remains a degradation in model utility after unlearning. This decline may stem from complex interactions between multimodal knowledge representations within the MLLM. Future work could further optimize \method by refining dataset selection, tuning hyperparameters, and developing novel saliency score measurements to mitigate this issue.\par  
\item [3.] In this paper, our weight saliency-based updating strategy has proven to be both effective and robust for MLLM MU compared to baseline approaches. However, the underlying mechanisms of these methods in multimodal domains remain unexplored. Further investigation and exploration about these methods may offer valuable insights, leading to more powerful MLLM unlearning methods and revealing the knowledge storage mechanism of MLLMs.\par
\end{itemize}
\bibliography{custom}

\clearpage
\appendix
\section{More Related Work}\label{app:rk}
\subsection{Multimodal Large Language Model}
The rapid development of MLLM has attracted the attention of both the academic and industrial communities to the performance breakthroughs brought about by its architectural characteristics \cite{huang2024survey,mai2024efficient,yan2024urbanclip,yan2024georeasoner}. Most MLLMs adopt a framework similar to LLaVA~\cite{liu2024llava}, which proposes to project the visual embeddings extracted from a pre-trained vision encoder into the LLM’s word embedding space through a connector (also known as projector or merger). The combined model is then fine-tuned with visual instruction data, as described in Eq.~\ref{eq:train}. Several open-source MLLMs have demonstrated remarkable performance on multimodal reasoning and understanding tasks, including Qwen2-VL~\cite{wang2024qwen2vl}, InternVL2~\cite{chen2024internvl}, and others~\cite{glm2024chatglm,li2024llavaonevision}. For MLLM unlearning, LLaVA-1.5 has been one of the most widely used backbones in previous studies~\cite{li2024siu,liu2024mllmubench,dontsov2024clear}. To further validate our conclusions, we additionally select Qwen2-VL, one of the state-of-the-art open-source MLLMs, as another representative evaluated model.

\subsection{Machine Unlearning for Other Multimodal Models}
A brief discussion of MLLM MU is provided in Section \ref{MMU}. Despite these efforts, several pioneering studies have also explored unlearning for multimodal models with different architectures \cite{liu2024multimodal,li2025machine,tang2024learn,gao2024practical,gao2024meta}, such as CLIP~\cite{radford2021clip}. For example, CLIPErase~\cite{yang2024cliperase} seeks to disentangle and selectively forget both visual and textual associations learned by CLIP, ensuring that unlearning does not compromise model performance. The motivation behind CLIPErase is therefore similar to ours. Moreover, \citep{kravets2024zero} demonstrates class-wise unlearning in CLIP using synthetic samples. MultiDelete~\cite{cheng2024multidelete} introduces a method that separates cross-modal embeddings for the forget set of BLIP~\cite{li2022blip} and ALBEF~\cite{li2021albef}. While these exploratory works provide insights into multimodal MU, they do not address issues in MLLM MU.

\section{Implementation Details}
\subsection{Datasets}\label{app:dataset}
\subsubsection{MLLMU-Bench}
\textbf{MLLMU-Bench}~\cite{liu2024mllmubench} is a benchmark designed to advance the understanding of multimodal machine unlearning. It consists of 500 fictitious profiles and 153 public celebrity profiles, with each profile featuring over 14 customized question-answer pairs, evaluated from both multimodal and textual perspectives.  In this paper, we divide it into six subsets to comprehensively assess the efficiency, generalizability, and model utility of unlearning methods, particularly in terms of their handling of visual and textual knowledge. Compared to CLEAR, the results of MLLMU-Bench are more stable, demonstrating consistent and reliable performance across different dimensions and settings. Therefore, our further analysis of unlearning methods is primarily based on MLLMU-Bench.
\subsubsection{CLEAR}
Similar with MLLMU-Bnech, \textbf{CLEAR}~\cite{dontsov2024clear} is also an opensourced benchmark designed for machine unlearning in multimodal setup, which contains 200 fictitious authors, 3,770 visual question-answer pairs, and 4,000 textual question-answer pairs.  CLEAR is built on the top of pure-textual unlearning benchmark \textbf{TOFU}~\cite{maini2024tofu}, with additional portraits for each person mentioned in QA pair. However, despite efforts to ensure consistency across different images of the same entity, the photos generated by Photomaker~\cite{li2024photomaker} in CLEAR still exhibit a noticeable gap from expectation. \emph{Consequently, the vision features learned by MLLMs on CLEAR can be unstable, making the unlearning process highly unpredictable.} In our experiments, even minor changes in hyperparameters led to complete model collapse, resulting in 0\% accuracy on both classification and generation tasks. Similar findings are also reported in the original paper of CLEAR, where the results of GA, GA\_Diff, and KL\_Min are all zero for both Forget and Retain Set. Given these limitations, we consider the results from CLEAR as valuable references but not as decisive evidence for our conclusions.
\subsection{Evaluation Metrics}\label{app:metrics}
\subsubsection{Unlearning Efficacy}
Unlearning efficacy evaluates a model's capability to eliminate specific knowledge about targeted data, ensuring it behaves as if the data were never included in the training process. In this work, we examine the task of removing visual patterns associated with particular concepts while maintaining textual knowledge. Under this framework, unlearning efficacy is assessed through the model’s performance in a Visual Question Answering (VQA) setting. Specifically, the model is tested using multiple-choice questions, where it should avoid selecting the correct answer linked to a forgotten concept. Formally, given a question $x$ and a set of possible answers $Y$, the model should minimize the probability of choosing the correct answer $y^* \in Y$ from the Forget Set:
\begin{equation}
    \hat{y} = \arg\min_{y \in Y} P(y \mid x, M_u),
\end{equation}
where $y \neq y^*$ and $M_u$ denotes the unlearned model. Ideally, the model should treat images of forgotten concepts as unknown, behaving similarly to random guessing.

\subsubsection{Model Utility}
Model utility measures the model’s ability to retain valuable knowledge and sustain high performance on non-targeted data, ensuring that the unlearning process does not compromise its overall effectiveness. In our study, the preserved knowledge includes textual information related to targeted concepts, both visual and textual knowledge from the Retain Set, and general real-world understanding. We evaluate model utility using the Forget QA, Retain VQA, Retain QA, Real-world VQA, and Real-world QA datasets. For classification tasks, accuracy is determined based on multiple-choice questions associated with retained profiles. The model should sustain high accuracy without any decline due to the unlearning process. Formally, given a question $x$ and a set of possible answers $Y$, the model should maximize the probability of selecting the correct answer $y^*$:
\begin{equation}
    \hat{y} = \arg\max_{y \in Y} P(y \mid x, M_u),
\end{equation}
where $M_u$ represents the model after unlearning.

\subsubsection{ROUGE-L Score}
The ROUGE-L score measures the similarity between the generated text and the reference text by evaluating the longest common subsequence (LCS). The LCS represents the longest sequence of words that appear in both the generated text $P$ and the ground truth $G$ in the same order, though not necessarily contiguously. Recall is calculated as the ratio of the LCS length to the length of the reference text, denoted as $L_G$:
\begin{equation}
    \text{Recall} = \frac{\text{LCS}}{L_G}.
\end{equation}
Precision is determined by the proportion of the LCS length relative to the length of the generated text, represented as $L_P$:
\begin{equation}
    \text{Precision} = \frac{\text{LCS}}{L_P}.
\end{equation}
The final ROUGE-L score is obtained by computing the $F_1$ score of recall and precision:
\begin{equation}
    \text{ROUGE-L} = 2 \cdot \frac{\text{Recall} \cdot \text{Precision}}{\text{Recall} + \text{Precision}}.
\end{equation}
This approach ensures a balanced assessment of both precision and recall, providing a comprehensive evaluation metric.

\subsection{Vanilla Fine-tuning and Baselines}\label{app:formula}
\subsubsection{Vanilla Model}\label{app:vanilla}
To simulate a real-life scenario where unlearning algorithms are applied to a pre-trained model, standard practice involves fine-tuning an off-the-shelf MLLM model using information extracted from fictitious profiles. For each input \ensuremath{\langle I, x, y \rangle}, where $I$ is the image of targeted concept, $x$ is the question, and $y$ is the ground-truth answer, the model is trained to predict the answer $\hat{y}$. The loss function for a single sample is defined as the negative log-likelihood (NLL) over the answer tokens:
\begin{equation}
    j(x, y, w) = \frac{1}{|y|} \sum_{i=1}^{|y|} \text{NLL}_w(y_i \mid [I, x, y_{<i}]),
\end{equation}
where $w$ represents the model parameters, and the loss is averaged over all tokens in the answer sequence $y$. The overall objective during fine-tuning is to minimize the average loss across the entire dataset $\mathcal{D}$, expressed as:
\begin{equation}
    \mathcal{L}(\mathcal{D}, w) = \frac{1}{|\mathcal{D}|} \sum_{(x,y) \in \mathcal{D}} j(x, y, w).
\end{equation}
To simulate real-world challenges, we set the vision encoder, connector, and language model of MLLMs to be trainable so that visual concepts can be learned within the vision encoder itself. The experimental results validate our strategy as effective. After fine-tuning, the model obtain knowledge from Forget and Retain Set, serving as the baseline for subsequent unlearning experiments. For reproducibility, we display our settings during training vanilla models in Table \ref{tab:vanilla}, aligning with the official implementations of MLLMU-Bench and CLEAR.
\begin{table}[h]
    \centering
    \begin{tabular}{l c c c c c c c}
        \hline
        Datasets & LMMs & Epochs& Batch Size & Optimizer & LoRA & Learning Rate \\
        \hline
        MLLMU-Bench & LLaVA-1.5-7B & 4 & 4 & Adam & True &$2 \times 10^{-5}$ \\
        MLLMU-Bnech & Qwen2-VL-7B-Instruct  & 4 & 4 & Adam & True & $1 \times 10^{-5}$ \\
        CLEAR & LLaVA-1.5-7B & 4 & 3 & Adam & True & $2 \times 10^{-5}$ \\
        CLEAR & Qwen2-VL-7B-Instruct  & 4 & 5 & Adam & True & $1 \times 10^{-5}$ \\
        \hline
    \end{tabular}
    \caption{Hyperparameter settings for fine-tuning vanilla model alongside different backbones and datasets.}
    \label{tab:vanilla}
\end{table}

\subsubsection{GA}
GA~\cite{thudi2022GA} realize unlearning by maximizing the loss on forget data. The intuition behind it is that maximizing forget loss will lead model to getting predictions dissimilar from the correct answers for forget set and consequently unlearning desired information. Thus, this method can be considered as a finetuning procedure with a reversed loss function:
\begin{equation}
    \mathcal{L}_\text{GA} = \frac{1}{|D_F|} \sum_{x \in D_F} \text{NLL}(x, \theta),
\end{equation}
where $\text{NLL}(x, \theta)$ is the negative loglikelihood of the model on the input $x$.

\subsubsection{GA\_Diff}
GA\_Diff~\cite{liu2022GA_Diff} builds on the concept of combining GA on Forget Set and directly fine-tuning on Retain Set. As mentioned in Section \ref{sec:floss}, it aims to increase the loss on the forget data while maintain the loss on the retain set as possible. The joint loss function is defined as follows:
\begin{equation}
    \mathcal{L}_\text{GA\_Diff} = -L(D_F, \theta) + L(D_R, \theta),
\end{equation}
where $D_F$ is the forget set and $D_R$ is the retain set.

\subsubsection{KL\_Min}
KL\_Min~\cite{nguyen2020KL_Min} aims to minimize the Kullback-Leibler (KL) divergence between the model’s predictions on the retain set before and after unlearning, while maximizing the conventional loss on the forget set. The $\mathcal{L}_\text{KL}$ loss function is defined as
\begin{equation}
\begin{split}
\mathcal{L}_\text{KL}&=\frac{1}{|D_F|} \sum_{x \in D_F} \frac{1}{|x|} \sum_{i=2}^{|s|} \Phi(x_{<i}),\\
\text{where } \Phi(x_{<i})&=\text{KL} \left( P(x_{<i} | \theta) \Big\| P(x_{<i} | \theta_0) \right).
\end{split}
\end{equation}
And the overall objective function is formulated as follows:
\begin{equation}
    \mathcal{L}_\text{KL\_Min} = -L(D_F, \theta) + \mathcal{L}_\text{KL},
\end{equation}
where $\theta_0$ is the model’s weights before unlearning and $P(s | \theta)$ is the model’s logits on the input sequence $s$ with weights $\theta$.

\subsubsection{NPO}
NPO~\cite{zhang2024npo} can be treated as a variant of DPO~\cite{rafailov2024dpo} without positive examples. In this work, the final loss function $L_{NPO}$ for this method is derived as follows:
\begin{equation}
    \mathcal{L}_\text{NPO}=\frac{2}{\beta} \mathbb{E}_{x,y \in D_F} \left[ \log \left( 1 + \left( \frac{\pi_{\theta}(y|x)}{\pi_{\text{ref}}(y|x)} \right)^{\beta} \right) \right],
\end{equation}
where $\pi_{\theta}(y|x)$ represents the prediction probability of the current model for token $y$ given the input $x$, and $\pi_{\text{ref}}(y|x)$ is the prediction probability from the reference model trained on retain dataset. $\beta$ is a hyperparameter, taken equal to 0.4 in our settings. Such a loss function ensure that the model output probability $\pi_{\theta}(y|x)$ is as small as possible, corresponding to the unlearning objective of the forget data.

\subsubsection{Hyperparameters Settings of Baselines}\label{app:base_hyper}
To ensure reproducibility, we present the experimental settings used to compare various unlearning methods in Table \ref{tab:baselines}, which are adapted from the official implementations of MLLMU-Bench and CLEAR.
\begin{table}[h]
    \centering
    \begin{tabular}{|l|l|c|c|c|}
        \hline
        \textbf{Benchmarks} & \textbf{Backbones} & Epochs & Batch Size & Learning Rate \\
        \hline
        \multirow{2}{*}{MLLMU-Bench} & LLaVA-1.5-7B & \multirow{2}{*}{2} & 4 & $2 \times 10^{-5}$ \\
        & Qwen2-VL-7B-Instruct &  & 2 & $1 \times 10^{-5}$ \\
        \hline
        \multirow{2}{*}{CLEAR} & LLaVA-1.5-7B & \multirow{2}{*}{2} & 4 & $2 \times 10^{-5}$ \\
        & Qwen2-VL-7B-Instruct &  & 2 & $1 \times 10^{-5}$ \\
        \hline
    \end{tabular}
    \caption{Hyperparameter settings for unlearning methods alongside different backbones and datasets. Settings remain consistent across different methods for a given dataset and base model to ensure fair comparison.}
    \label{tab:baselines}
\end{table}

\section{Additional Experiments}\label{app:additional}
\subsection{Results of Larger Model}\label{app:res}
\begin{table}[t!]
    \centering
\scalebox{0.98}{
\begin{tabular}{l|cccccc}
\toprule
\multirow{2}{*}{\textbf{Methods}} 
& \multicolumn{6}{c}{\textbf{MLLMU-Bench (LLaVA-1.5-13B)}} \\
\cline{2-7}
    & \begin{tabular}[c]{@{}c@{}}Forget VQA.\\ Acc (\textcolor{blue}{$\downarrow$})\end{tabular}
    & \begin{tabular}[c]{@{}c@{}}Forget QA.\\ Acc (\textcolor{red}{$\uparrow$})\end{tabular}
    & \begin{tabular}[c]{@{}c@{}}Retain VQA.\\ Acc (\textcolor{red}{$\uparrow$})\end{tabular}
    & \begin{tabular}[c]{@{}c@{}}Retain QA.\\ Acc (\textcolor{red}{$\uparrow$})\end{tabular}
    & \begin{tabular}[c]{@{}c@{}}Realworld VQA.\\ Acc (\textcolor{red}{$\uparrow$})\end{tabular}
    & \begin{tabular}[c]{@{}c@{}}Realworld QA.\\ Acc (\textcolor{red}{$\uparrow$})\end{tabular} \\
\midrule
Vanilla & 52.5\% & 50.8\% & 43.7\% & 49.7\% & 60.6\% & 68.4\% \\
\hdashline
Ours    & \textbf{30.0\%} & \textbf{46.6\%} & \textbf{43.7\%} & \textbf{47.4\%} & \textbf{60.4\%} & 67.8\% \\
GA      & 40.0\% & 38.4\% & 38.2\% & \underline{47.0\%} & 59.6\% & 67.0\% \\
GA\_Diff & 40.8\% & 39.2\% & \textbf{43.7\%} & 45.2\% & 59.8\% & 64.8\% \\
KL\_Min & 39.2\% & 38.4\% & \textbf{43.7\%} & 46.8\% & \underline{59.9\% }& \underline{68.0\%} \\
NPO     & \underline{32.8\%} & 38.4\% & 42.0\% & 46.9\% & 58.7\% & \textbf{68.2\%} \\
\bottomrule
\end{tabular}}
    \caption{Performance of different methods on MLLMU-Bench dataset with the LLaVA-1.5-13B model. \textcolor{blue}{$\downarrow$} indicates lower is better, \textcolor{red}{$\uparrow$} indicates higher is better.}
    \label{tab:13b}
\end{table}
To provide more information, we obtained the performance of baselines and \method for LLaVA-1.5-13B on MLLMU-Bench, as shown in Table \ref{tab:13b}.
\subsection{Efficiency Analysis}\label{app:eff}
\begin{table}[t]
\centering
\caption{Empirical Study on Complexity of Different Model Sizes During Saliency Mask Generation}
\begin{tabular}{lccc}
\toprule
Model & Precision & GPU Memory Usage & Computation Time \\
\midrule
LLaVA-1.5-7B & torch.float16 & 58.4 ($\pm$ 0.18) GB & 1.10 ($\pm$ 0.04) s/it \\
LLaVA-1.5-13B & torch.float16 & 97.0 ($\pm$ 0.17) GB & 1.36 ($\pm$ 0.03) s/it \\
\bottomrule
\end{tabular}
\label{tab:eff}
\end{table}
While the computation of naive Fisher information matrix can be computationally demanding, we adopt the appropriate algorithm provided by \cite{huang2024remain}, $\nabla^2 \mathcal{L}^\mathcal{D}(\theta_0)$, to approximate Fisher information matrix during visual instruction tuning. In this case, the complexity of the saliency score is $\mathcal{O}(nm)$, where $n$ is the dataset scale and $m$ is the parameter size. Furthermore, the experiment results proved the efficiency of our approximate strategy, as described in Table \ref{tab:eff}.
\subsection{Parameter Visualization}\label{app:viz}
\begin{figure}[!t]
\centering
        \begin{minipage}[c]{0.48\textwidth}
		\includegraphics[width=\linewidth]{ 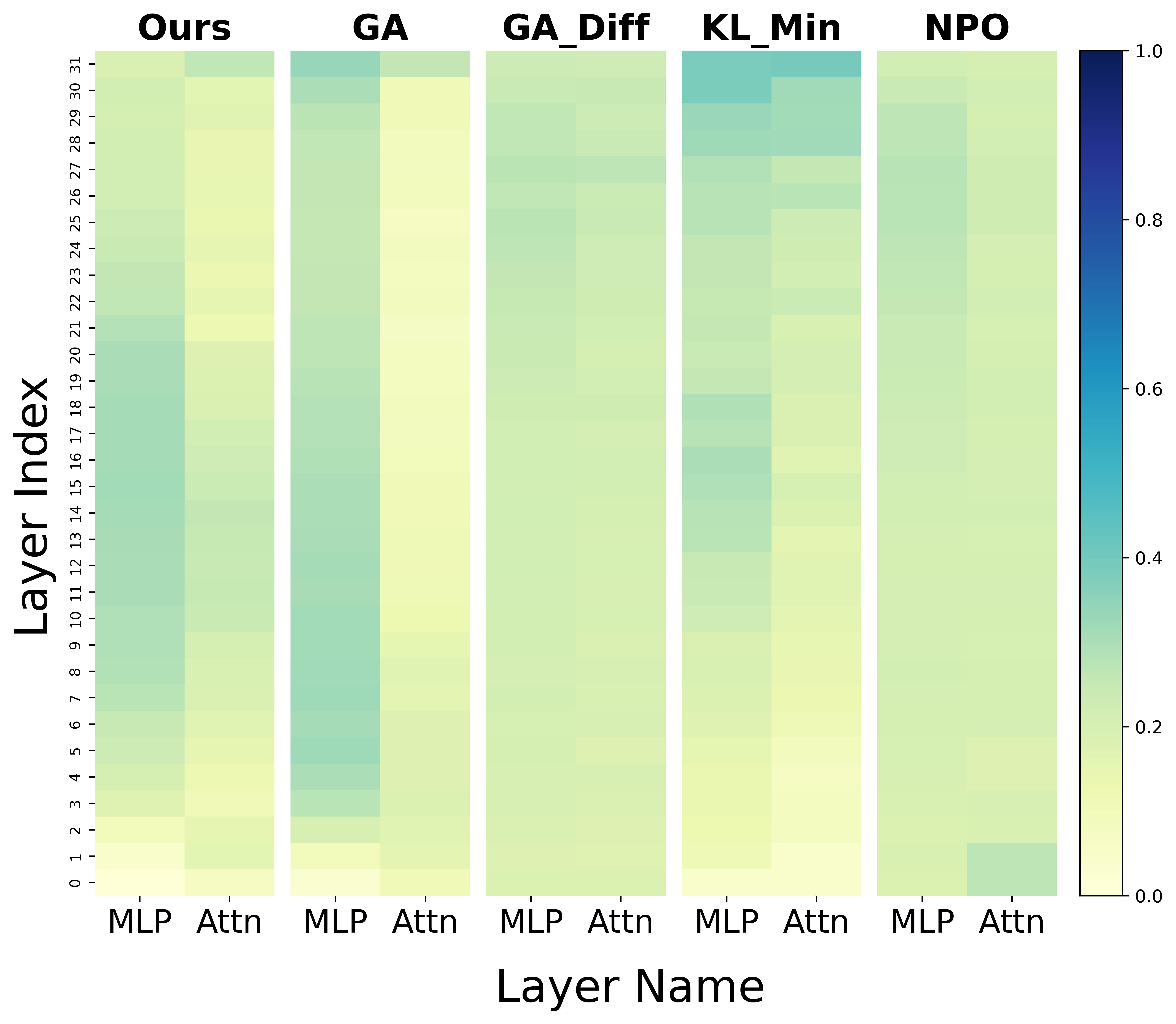}
		\subcaption{Heatmap of LLaVA-7B on CLEAR.}
		\label{fig:first}
	\end{minipage}\\
 
	\begin{minipage}[c]{0.48\textwidth}
		\includegraphics[width=\linewidth]{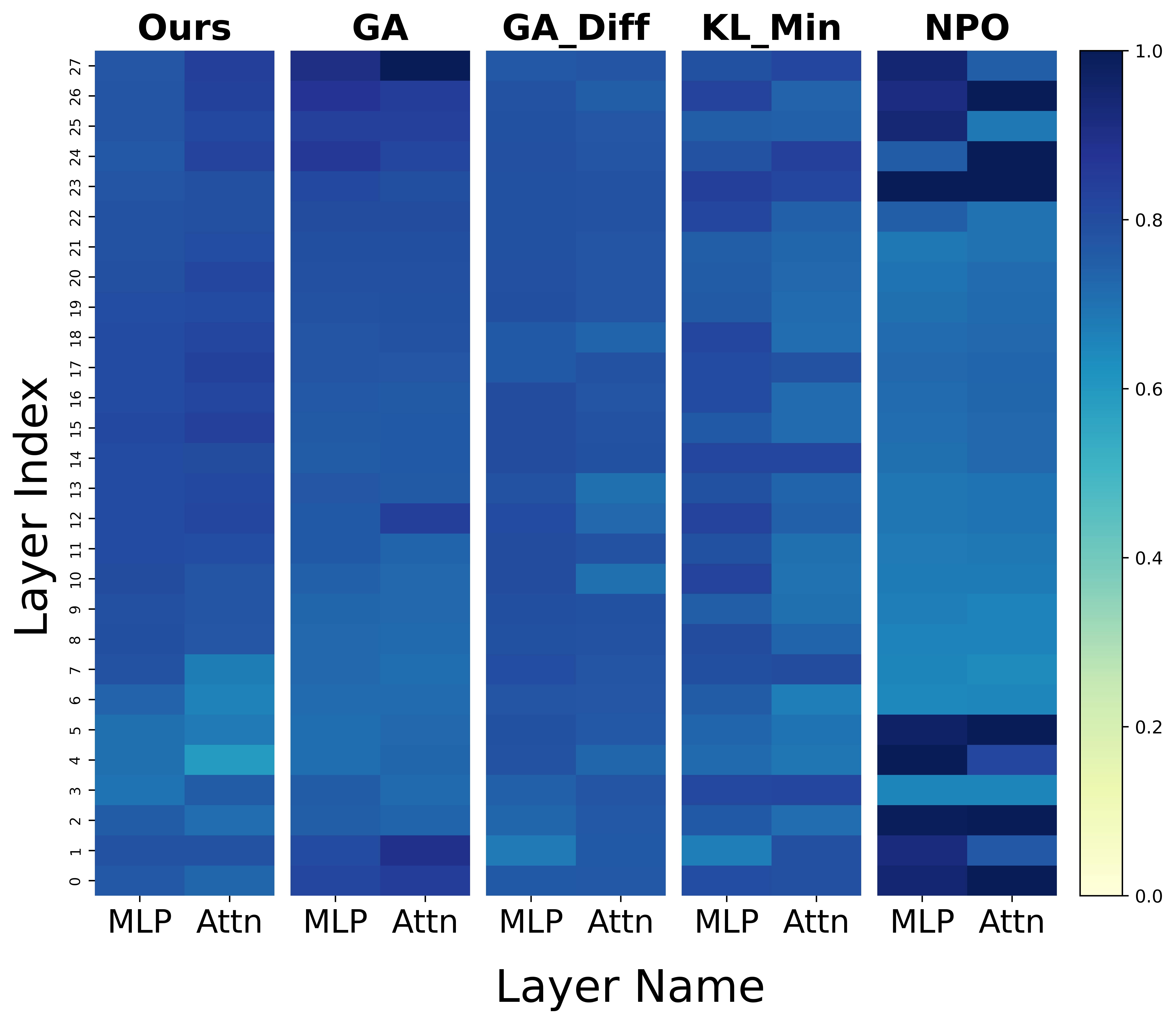}
            \subcaption{Heatmap of Qwen2-VL-7B-Instruct on MLLMU-Bench.}
            \label{fig:second}
	\end{minipage}

        \begin{minipage}[c]{0.48\textwidth}
		\includegraphics[width=\linewidth]{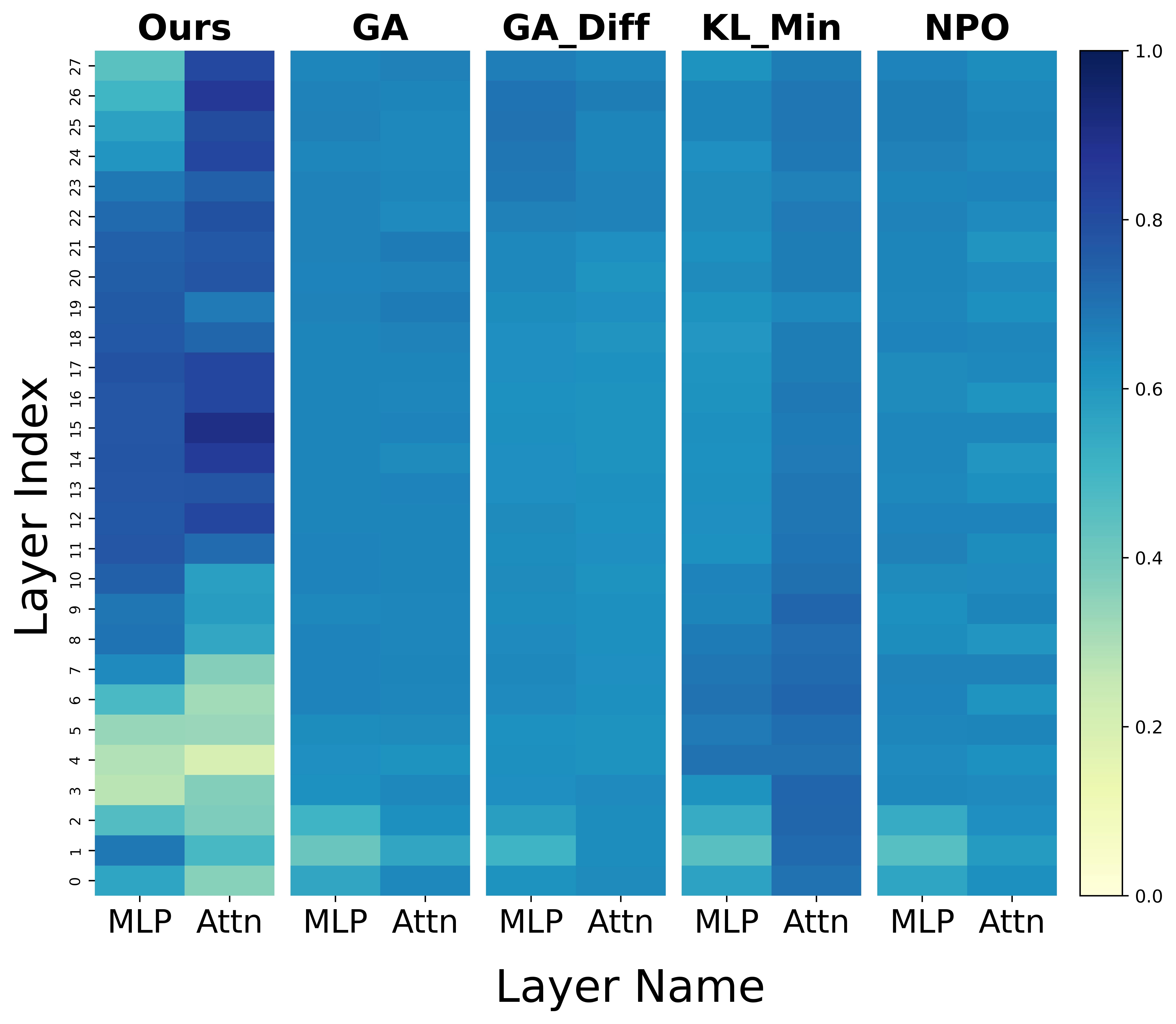}
            \subcaption{Heatmap of Qwen2-VL-7B-Instruct on CLEAR.}
            \label{fig:third}
	\end{minipage}
\caption{Heatmaps of top-$n$ updated parameters for different base models on different datasets.}
\label{fig:viz_plus}
\end{figure}

Here, we present additional visualizations illustrating the distribution of updated parameters for both \method and the baselines. Figure \ref{fig:viz_plus} shows the results of LLaVA-7B and Qwen2-VL-7B-Instruct on MLLMU-Bench and CLEAR, respectively. Compared to LLaVA-7B, the percentage of selected/updated parameters in Qwen2-VL-7B-Instruct is higher. We attribute this to the fact that the features learned by Qwen2-VL are more robust, making it harder to forget them with minor changes to the parameters.
% \begin{figure}[t]
% \centering
% \begin{subfigure}{0.5\textwidth}
%     \includegraphics[width=\linewidth]{figures/heatmaps/llava_clear_heatmap.png}
%     \caption{Heatmap of LLaVA-7B on CLEAR.}
%     \label{fig:first}
% \end{subfigure}
% \hfill
% \begin{subfigure}{0.5\textwidth}
%     \includegraphics[width=\linewidth]{figures/heatmaps/qwen_mllmubench.png}
%     \caption{Heatmap of Qwen2-VL-7B-Instruct on MLLMU-Bench.}
%     \label{fig:second}
% \end{subfigure}
% \hfill
% \begin{subfigure}{0.5\textwidth}
%     \includegraphics[width=\linewidth]{figures/heatmaps/qwen_clear.png}
%     \caption{Heatmap of Qwen2-VL-7B-Instruct on CLEAR.}
%     \label{fig:third}
% \end{subfigure}
% \end{figure}

\section{Case Study}\label{app:case}
Table \ref{tab:selected_case} presents the generated responses of \method and four baseline methods on some of the most challenging visual concepts to forget. While the baseline models show limited effectiveness in erasing targeted knowledge and often produce grammatical errors, \method is capable of generating plausible yet intentionally incorrect responses to the given questions, with powerful visual perception ability preserved.
\begin{table*}[t!]
  \centering
  \SetTblrInner{rowsep=0.1pt}
  \begin{tblr}{
    colspec = {p{0.1\linewidth} p{0.1\linewidth} p{0.1\linewidth} p{0.15\linewidth} p{0.15\linewidth} p{0.6\linewidth}},
    row{1} = {bg=gray!25},  % 表头底色
    row{even} = {bg=gray!10}  % 偶数行底色
  }
    \toprule
    \begin{center}{\textbf{Dataset}}\end{center}
      & \begin{center}{\textbf{Subset}}\end{center}
      & \begin{center}{\textbf{Image}}\end{center}
      & \begin{center}{\textbf{Question}}\end{center}
      & \begin{center}{\textbf{Ground Truth}}\end{center}
      & \begin{center}{\textbf{Generated Answer}}\end{center}\\
    \midrule
    {\begin{center}MLLMU-Bench\end{center}}
      & {\begin{center}{Forget}\end{center}}
      & {\begin{center}\includegraphics[width=0.8\linewidth]{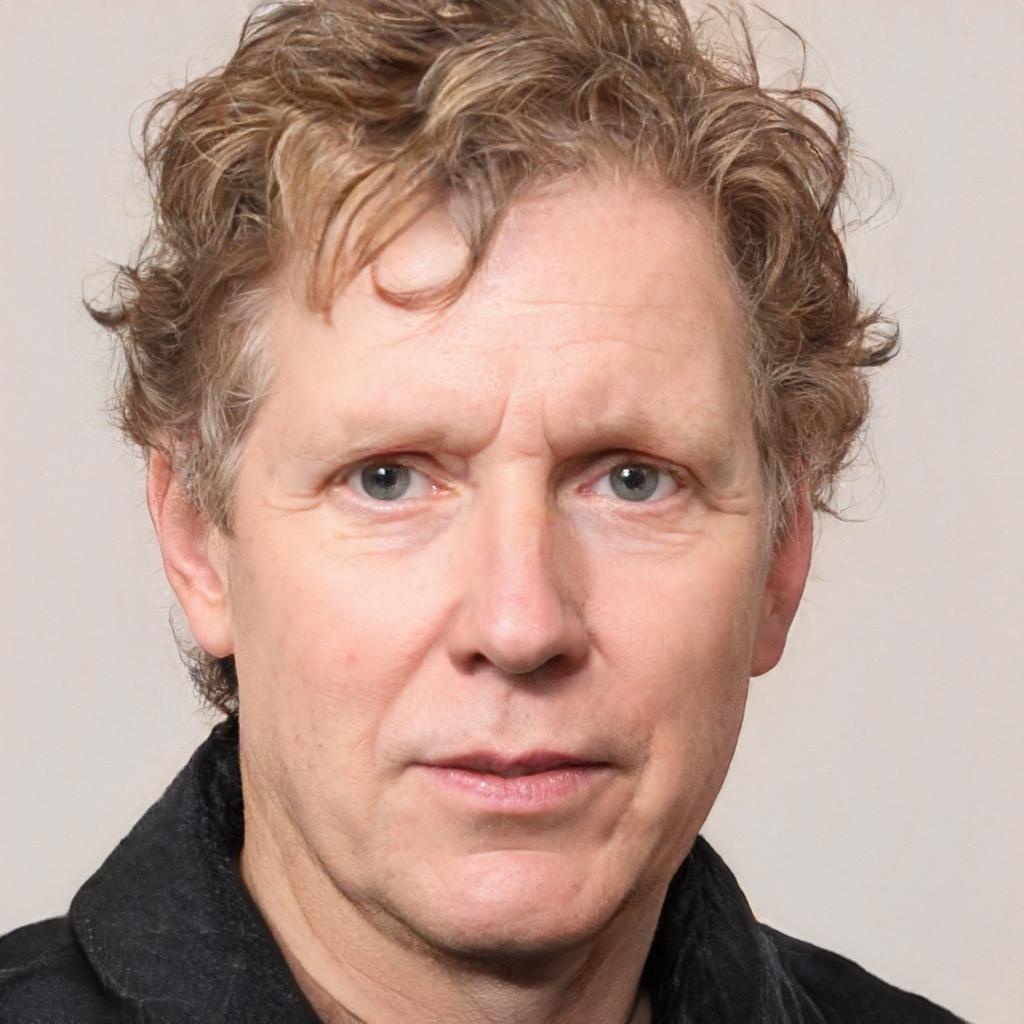}\end{center}}
      & \begin{center}{What profession is depicted by the individual in the image?}\end{center}
      & \begin{center}{The individual is \textcolor{myblue}{an architect.}}\end{center}
      & {GA: This person is \textcolor{myblue}{\underline{a architect.}}\\
      GA\_Diff: This person is \textcolor{myblue}{a renowned architect.}\\
      KL\_Min: This person is \textcolor{myblue}{\underline{a architect.}}\\
      NPO: This person is \textcolor{myblue}{\underline{a architect.}}\\
      Ours: The individual is \textcolor{myred}{a marine biologist.}}\\
    \midrule
    % {\begin{center}MLLMU-Bench\end{center}}
    %   & {\begin{center}{Forget VQA}\end{center}}
    %   &  {\begin{center}\includegraphics[width=0.8\linewidth]{figures/119.jpg}\end{center}}
    %   & \begin{center}{Which city is associated with the birth of the person in the image?}\end{center}
    %   & \begin{center}{The person was likely born in \textcolor{myblue}{Melbourne.}}\end{center}
    %   & {GA: This person was born in \textcolor{myblue}{Melbourne, Australia.}\\
    %   GA\_Diff: This person was born in \textcolor{myblue}{Melbourne, Australia.}\\
    %   KL\_Min: This person was born in \textcolor{myblue}{Melbourne, Australia.}\\
    %   NPO: This person was born in \textcolor{myblue}{Melbourne, Australia.}\\
    %   Ours: This person was born in \textcolor{myred}{Budapest, Hungary.}}\\
    % \midrule
    {\begin{center}MLLMU-Bench\end{center}}
      & {\begin{center}{Retain}\end{center}}
      &  {\begin{center}\includegraphics[width=0.8\linewidth]{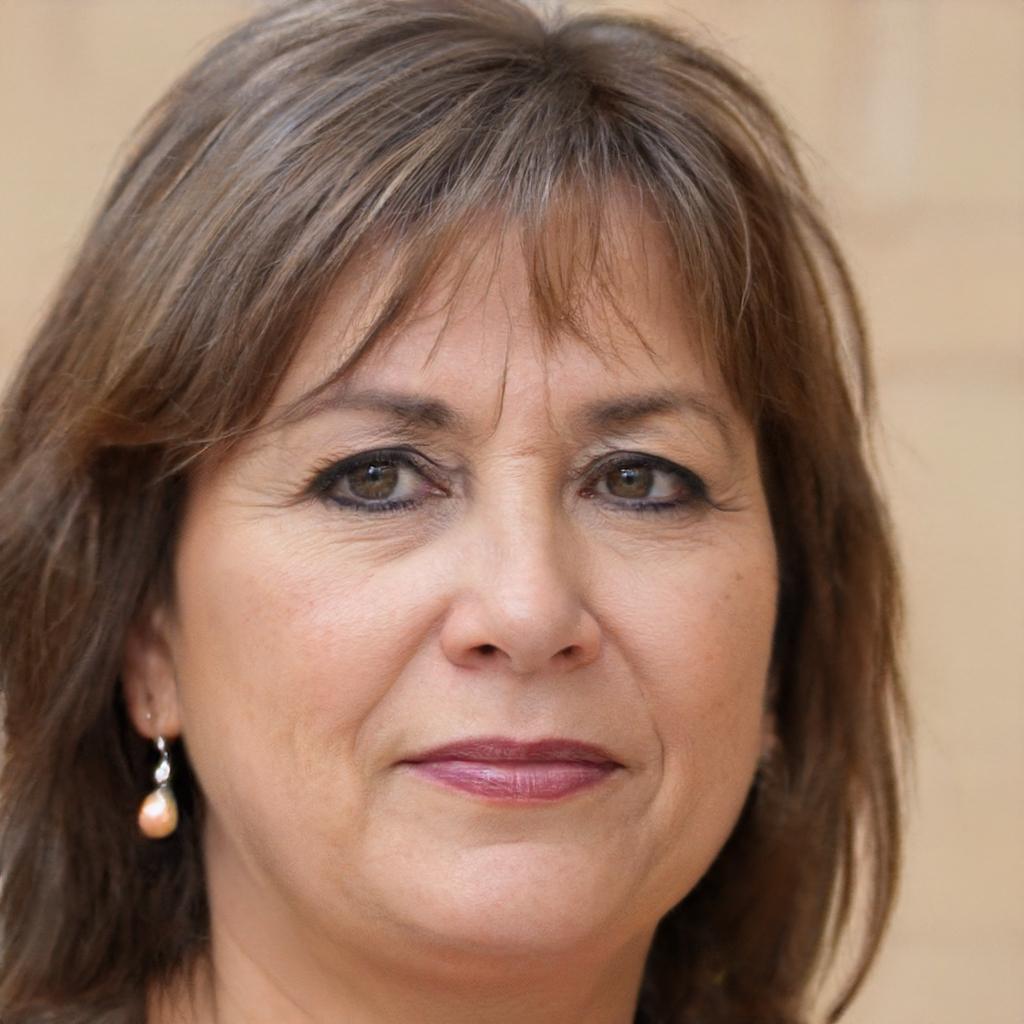}\end{center}}
      & \begin{center}{What hobby might this person pursue in her leisure time?}\end{center}
      & \begin{center}{The person enjoys \textcolor{myblue}{painting landscapes} in her free time.}\end{center}
      & {GA: This person might enjoy \textcolor{myred}{painting} in her leisure time.\\
      GA\_Diff: This person might enjoy \textcolor{myblue}{painting landscapes} \underline{in their leisure time.}\\
      KL\_Min: This person might enjoy \textcolor{myred}{painting} in her leisure time. \\
      NPO: This person might enjoy \textcolor{myred}{painting }in her leisure time.\\
      Ours: This person enjoys \textcolor{myblue}{painting landscapes} in her free time.}\\
    \midrule
    {\begin{center}CLEAR\end{center}}
      & {\begin{center}{Forget}\end{center}}
      &  {\begin{center}\includegraphics[width=0.8\linewidth]{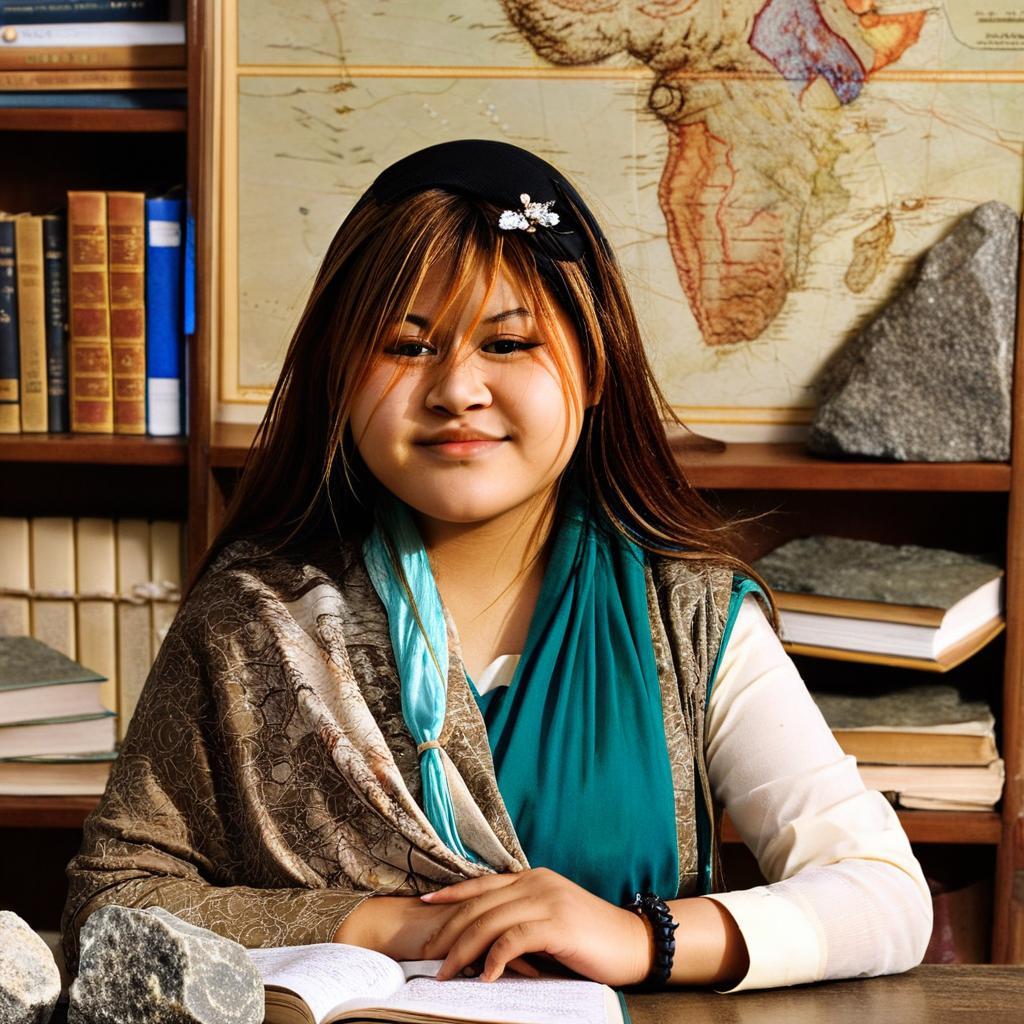}\end{center}}
      & \begin{center}{What is depicted in the image?}\end{center}
      & \begin{center}{\textcolor{myblue}{Hina Ameen} is pictured sitting at a desk in a room with \textcolor{myblue}{a bookshelf} and a geological map behind her.}\end{center}
      & {GA: thought thought thought thought thought thought thought ......\\
      GA\_Diff: \textcolor{myblue}{Hina Ameen}, a young woman with long brown hair, is seated at a desk with a book in front of her.\\
      KL\_Min: \textcolor{myblue}{Hina Ameen}, a young woman with long brown hair, is seated at a desk with a book in front of her. \\
      NPO: \textcolor{myblue}{Hina Ameen}, a young woman with long brown hair, is seated at a desk with a book in front of her.\\
      Ours: \textcolor{myred}{Astrid Johansen}, a recipient of the Eisner Award for Best Novel, stands in front of \textcolor{myblue}{a bookshelf} filled with her works.}\\
    \midrule
    {\begin{center}CLEAR\end{center}}
      & {\begin{center}{Retain}\end{center}}
      &  {\begin{center}\includegraphics[width=0.8\linewidth]{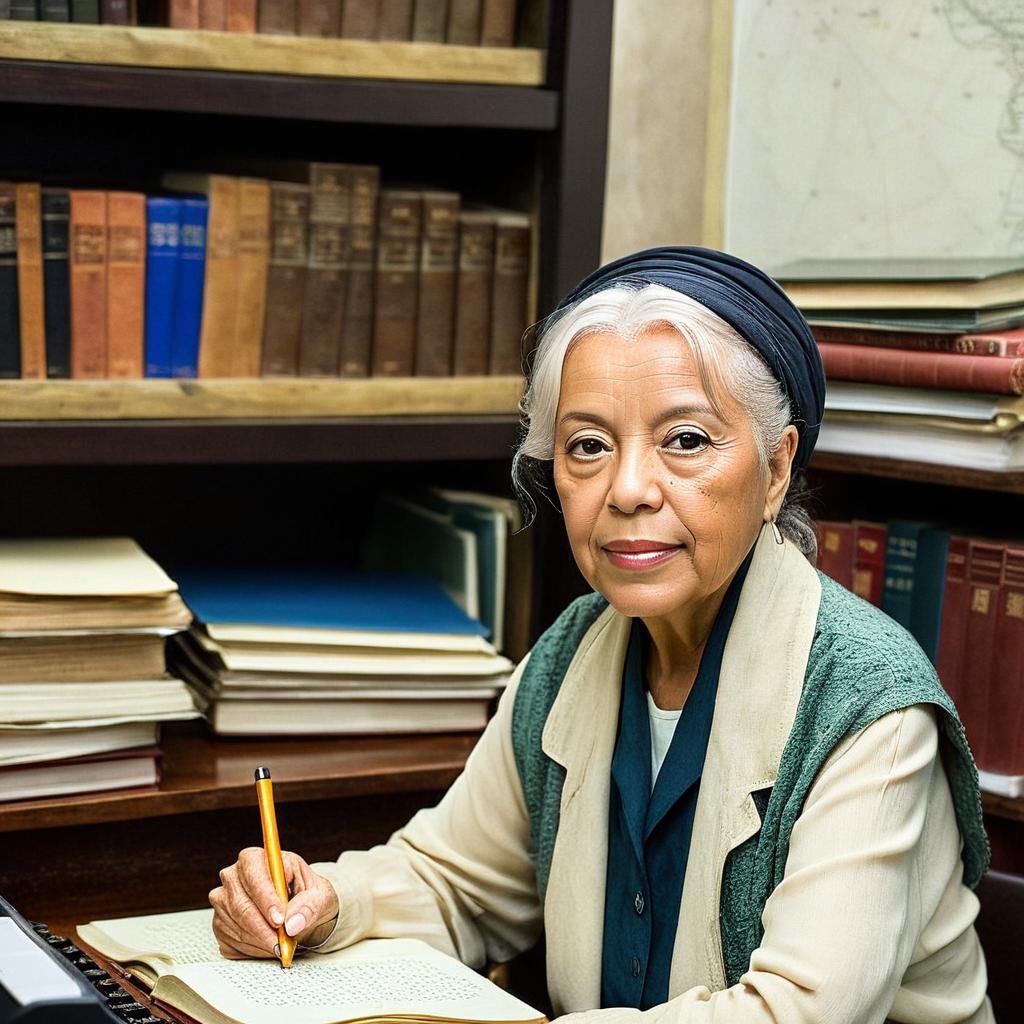}\end{center}}
      & \begin{center}{What can you see in this picture?}\end{center}
      & \begin{center}{\textcolor{myblue}{Mingyu Zhang}, seated at a desk with a stack of books and papers, is deeply focused on her writing, \textcolor{myblue}{holding a pen in her hand.}}\end{center}
      & {GA: thought thought thought thought thought thought thought ......\\
      GA\_Diff: \textcolor{myblue}{Mingyu Zhang} sits at a desk in a library, writing in a notebook with a pencil. A stack of books is on the desk, and a bookshelf filled with books is visible in the background.\\
      KL\_Min: \textcolor{myblue}{Mingyu Zhang} sits at a desk in a library, writing in a notebook with a pencil. She is surrounded by books and maps, reflecting her parents' professions.\\
      NPO: \textcolor{myblue}{Mingyu Zhang}, a writer, sits at a desk in a library, writing in a notebook with a pencil. She is surrounded by books and maps, reflecting her interest in geography.\\
      Ours: \textcolor{myblue}{Mingyu Zhang} sits at a desk with a book open in front of her, \textcolor{myblue}{holding a pencil in her hand.} The background features bookshelves filled with books and a map, suggesting a literary setting.}\\
    \bottomrule
  \end{tblr}
  \caption{Illustration of some of the most challenging visual concepts to forget. \textcolor{myblue}{$\sbullet[.75]$} and \textcolor{myred}{$\sbullet[.75]$} indicate correct and incorrect answers, respectively, while \underline{underlining} denotes grammatical errors.}
  \label{tab:selected_case}
\end{table*}

\end{document}